\documentclass[5p]{elsarticle}

\usepackage{lineno}
\usepackage{hyperref}
\usepackage{amsmath}
\usepackage{xcolor}
\usepackage{layouts}
\usepackage{graphicx}
\usepackage{adjustbox}
\usepackage{booktabs}
\usepackage{subfig}

\usepackage[normalem]{ulem}
\usepackage{color}

\newcommand{\rem}[1]{}
\newcommand{\add}[1]{#1}
\newcommand{\addtwo}[1]{#1}

\biboptions{authoryear}

\makeatletter
\def\ps@pprintTitle{%
 \let\@oddhead\@empty
 \let\@evenhead\@empty
 \def\@oddfoot{\textit{\hfill\today}}%
 \let\@evenfoot\@oddfoot}
\makeatother

\begin{document}
\setlength{\emergencystretch}{3em}
\begin{frontmatter}

\title{\addtwo{Global canopy height regression and uncertainty estimation from GEDI LIDAR waveforms with deep ensembles}}

\author[1]{Nico Lang\corref{cor1}}
\author[1]{Nikolai Kalischek}
\author[2]{John Armston}
\author[1]{Konrad Schindler}
\author[2]{Ralph Dubayah}
\author[1,3]{Jan Dirk Wegner}

\address[1]{EcoVision Lab, Photogrammetry and Remote Sensing, ETH Zürich, Switzerland}
\address[2]{Department of Geographical Sciences, University of Maryland, College Park, MD, USA}
\address[3]{Institute for Computational Science, University of Zurich, Switzerland}
\cortext[cor1]{Corresponding author: nico.lang@geod.baug.ethz.ch}

\begin{abstract}
NASA’s Global Ecosystem Dynamics Investigation (GEDI) is a key climate mission whose goal is to advance our understanding of the role of forests in the global carbon cycle. While GEDI is the first space-based LIDAR explicitly optimized to measure vertical forest structure predictive of aboveground biomass, the accurate interpretation of this vast amount of waveform data across the broad range of observational and environmental conditions is challenging. Here, we present a novel supervised machine learning approach to interpret GEDI waveforms and regress canopy top height globally. We propose a \addtwo{probabilistic deep learning approach based on an ensemble of deep convolutional neural networks (CNN)}\rem{Bayesian convolutional neural network} to avoid the explicit modelling of unknown effects, such as atmospheric noise. The model learns to extract robust features that generalize to unseen geographical regions and, in addition, yields reliable estimates of predictive uncertainty. Ultimately, the global canopy top height estimates produced by our model have an expected RMSE of 2.7 m with low bias. 
\end{abstract}


\end{frontmatter}


\section{Introduction}

Forests play a key role in the carbon cycle and in the regulation of the climate~\citep{mitchard2018tropical}.  
However, estimates of carbon \add{fluxes} between the atmosphere and the terrestrial biosphere suffer from high uncertainty, with a relative standard deviation of up to 47\%~\citep{friedlingstein2019global}. \add{For the last decade (2009--2018), annual CO$_\text{2}$ emissions from global land-use change, mainly from deforestation, account for 1.5$\pm$0.7~GtC~yr$^{-1}$}, and the global terrestrial CO$_\text{2}$ sink amounts to 3.2$\pm$0.6~GtC~yr$^{-1}$~\citep{friedlingstein2019global}.
Although the aboveground biomass (hereafter "biomass") of forests is crucial to estimate terrestrial carbon flux, existing global biomass products differ significantly \add{in spatial patterns}~\citep{mitchard2013uncertainty, avitabile2016integrated, mitchard2018tropical, ploton2020spatial, spawn2020harmonized}. Tropical carbon stocks are particularly difficult to estimate \add{due to spatial heterogeneity and saturation of remote sensing signals}~\citep{kohler2010towards,mitchard2014markedly, phillips2014evaluating}, contributing to high uncertainty about the tropical land-use change flux~\citep{pan2011large}. This lack of accuracy in global biomass maps hinders our ability to assess the impact of deforestation and subsequent regrowth on atmospheric CO$_\text{2}$ concentrations, limiting the quality of climate simulations and forecasts~\citep{kunreuther2014integrated}. 
\add{In addition, better maps of aboveground biomass are needed to protect various ecosystem services, by optimising spatial planning and by identifying priority areas for conservation and restoration~\citep{strassburg2020global}.}

Canopy height is an important predictor for aboveground biomass~\citep{jucker2017allometric, dubayah2010estimation, qi2019forest,  carreiras2017coverage, asner2014mapping, silva2018comparison, drake2002estimation, drake2002sensitivity}.
It is also a key morphological trait which, along with composition and function, helps to \add{monitor ecosystem response to climate and land use change as well as restoration}~\citep{schneider2017mapping, valbuena2020standardizing}. Additionally, \add{ecosystem structure, which is related to canopy height and its heterogeneity, is} an essential biodiversity variable and a strong predictor for species richness at local to global scales~\citep{kreft2007global,gatti2017exploring,marselis2020evaluating}.
Previous pan-tropical and global biomass maps~\citep{baccini2012estimated,saatchi2011benchmark} were based on canopy height derived from space-borne LIDAR data collected by NASA's Ice, Cloud and Elevation Satellite (ICESat)~\citep{abshire2005geoscience}. ICESat was optimised to retrieve ice surface elevations and ceased operation in 2010. A new space-borne LIDAR, NASA's Global Ecosystem Dynamics Investigation (GEDI), was launched to the International Space Station (ISS) in late 2018 ~\citep{DUBAYAH2020100002}.
GEDI is the first space-borne, full waveform LIDAR that is specifically designed to measure ecosystem structure by providing vertical profiles of forest canopies \add{with 25~m footprints on the ground.} After its nominal mission duration of two years, the mission is expected to have sampled 4~\% of the land surface between 51.6 degree \add{north and south} on a sparse sampling grid. 

One of the major challenges of using space-borne LIDAR concerns the processing of its waveforms. Features such as the location of the ground and the top of the canopy must be identified to derive canopy height and other structure variables, such as canopy cover and vertical plant area profiles. The effects of known (e.g., pulse shape, digitizer noise) and unknown (e.g., slope, multiple scattering) properties are often difficult to explicitly model in the separation of canopy and ground signals and determination of ranging points along the waveform. For example, \add{under} dense canopy cover conditions, the ground return may be quite weak; detecting this weak signal against high background noise is difficult. Similarly, detecting the very tops of trees, where there may be little leaf area, and thus a weak return, will lead to errors in height recovery~\citep{DUBAYAH2020100002}. 

Traditional approaches to LIDAR waveform processing~\citep{hofton_GEDI_ATBD} have relied on conventional signal processing, but GEDI presents a challenge to these methods for several reasons. First, it will produce an unprecedented \add{high} number of fine spatial resolution observations. These span the huge range of topographic, climatic, and edaphic conditions across the Earth’s temperate and tropical forests, leading to tremendous structural variability of the waveforms. Secondly, GEDI utilizes both high and low power beams which, when coupled with the structural diversity, variations in noise levels, and potential instrument artifacts, complicate the calibration. The unprecedented amount of space-borne measurements from GEDI thus presents a challenge, but also an opportunity to explore new data-driven algorithms based on machine learning.

In this work, we propose one such data-driven approach, based on state-of-the-art \rem{Bayesian}\addtwo{probabilistic} deep neural networks, that estimates global canopy heights together with well-calibrated uncertainties from geolocated (Level 1B/L1B) GEDI waveforms~\citep{dubayah2020gedi_L1B}. 
Our approach increases the retrieval performance compared to initial GEDI canopy height products (Level 2A/L2A) that were processed with traditional methods, but using provisional calibrations~\citep{dubayah2020gedi_L2A}, reducing the root mean square error (RMSE) from 4.4~m to 2.7~m and the underestimation bias from $-1.0$~m to $-0.1$~m (mean error).
We parameterize this direct, functional mapping from GEDI L1B waveforms to canopy top heights with a one-dimensional convolutional neural network (CNN), with a \add{residual block} architecture~\citep[ResNet,][]{he2016deep} at its core. 
Our proposed CNN uses the full GEDI L1B waveform as its input and transforms it \add{into} representative features \add{(recognizing peaks and patterns in the waveform)} that are suitable for the regression of a waveform height metric (the 98\textsuperscript{th} percentile height, or RH98) as a proxy for the canopy top height. 

A key advantage of our \rem{Bayesian}\addtwo{probabilistic} approach is that the model directly provides a well-calibrated uncertainty together with every estimate. 
This property is crucial to model deployment at global scale, because reference data for validation are limited.
We model two sources of uncertainty: the \emph{aleatoric} and the \emph{epistemic} uncertainty~\citep{der2009aleatory,kendall2017uncertainties}. The former describes the stochasticity in the data (e.g., caused by constantly varying atmospheric transmission and surface reflectance, as well as sensor noise), which is inherent, and is not reduced by collecting more training data. 
This uncertainty is estimated by minimizing the Gaussian negative log likelihood, a common approach for regression tasks~\citep{gast2018lightweight}. In other words, instead of outputting a point prediction, the network predicts a distribution over the output, \add{which is parameterized by a mean and a variance,} to approximate the conditional distribution of the target canopy height given an input waveform.
The second source is the epistemic uncertainty \add{(also known as \emph{model} uncertainty)} that describes imperfect knowledge of the model parameters. In our supervised machine learning approach, this knowledge is extracted from the training data. In contrast to the aleatoric uncertainty, it can be reduced by collecting more samples to get a more complete representation of the data distribution. 
We estimate the epistemic uncertainty with an ensemble of ten separate CNNs that are trained independently, starting from different random initializations~\citep{lakshminarayanan2017simple}. The variance computed over their individual \add{mean} predictions yields an estimate of the epistemic uncertainty \add{and the aleatoric uncertainty is computed by averaging the individual variance predictions.}
The overall \textit{predictive uncertainty} assigned to each canopy height estimate is the sum of both sources. This predictive uncertainty plays an important role to inform downstream applications that build on the canopy height. Furthermore, accompanying canopy height estimates with uncertainties enables the user to directly filter them to reduce the overall error. For example, in our case, discarding the 30\% most uncertain  predictions reduces the error by 25\%, while preserving the full range of canopy heights.

To train and evaluate the proposed neural network, a dataset of 68,483 on-orbit GEDI waveforms were matched to the canopy top heights (so-called RH98, which is the height of the 98th percentile of the cumulative waveform energy relative to the ground) derived from airborne laser scanning (ALS) reference data. To reduce noise caused by geolocation errors of the first release GEDI data (mean horizontal offset $\approx$20~m), \add{blocks of GEDI on-orbit waveforms are optimally correlated to simulated waveforms derived from the ALS data~\citep{hancock2019gedi}}. This reduces geolocation errors to a few meters on average and yields a high quality training and validation dataset.
We examine the error characteristics of our CNN predictions regarding forest structure, topography, and laser power. To test if our model is globally applicable, we carry out a geographic cross-validation, where we train the model without any samples from a particular continental region and then test it on that held-out region. This avoids overly optimistic interpretations due to spatial auto-correlation~\citep{ploton2020spatial}.

We present global canopy height maps based on the first four months of L1B mission data to demonstrate that our approach may be applied across the vast range of conditions that exist at the global scale.
While this work focuses on the estimation of the RH98 as a proxy for the canopy top height, we also show that the proposed machine learning approach is generic and can be applied to estimate other forest structure variables and the ground elevation from on-orbit GEDI waveforms.

\add{\section{Data}
\subsection{GEDI L1B waveforms product}
The GEDI laser system is mounted onboard the international space station (ISS) flying at 415~km altitude and measures along eight ground tracks that are 600~m apart. 
\add{Four of the ground tracks are produced by two full-power lasers and the other four tracks by one so-called coverage laser with reduced power.} 
Individual laser shots are spaced 60~m along track and have an approximately 25~m footprint on the ground. First measurements were acquired in April 2019 and \add{after its nominal life time of two years, ten billion waveforms will be measured over the land surface at 25~m spatial resolution}~\citep{DUBAYAH2020100002}.
GEDI provides the L1B full waveform product~\citep{dubayah2020gedi_L1B} with a maximum waveform length of 1420 vertical bins and a fixed interval of 1~ns ($\sim$15~cm) between consecutive returns.
\subsection{GEDI L2A elevation and height metrics product}
The L2A GEDI mission product contains elevation and relative height metrics derived from the L1B waveforms~\citep{dubayah2020gedi_L2A}. The first release of the L2A product (R001) contains six different algorithm setting groups that are designed to minimize false positive errors in detecting the ground and canopy top under a wide range of environmental conditions.
In addition, the L2A product provides geographical locations of the lowest mode (ground) and the percentage of non-vegetated area from the MODIS product MOD44B Version 6 Vegetation Continuous Fields~\citep{dimiceli2015mod44b}.}
\subsection{Airborne LIDAR reference dataset}

\begin{figure*}[]
    \centering
    
    \subfloat[]{{\includegraphics[width=.98\textwidth]{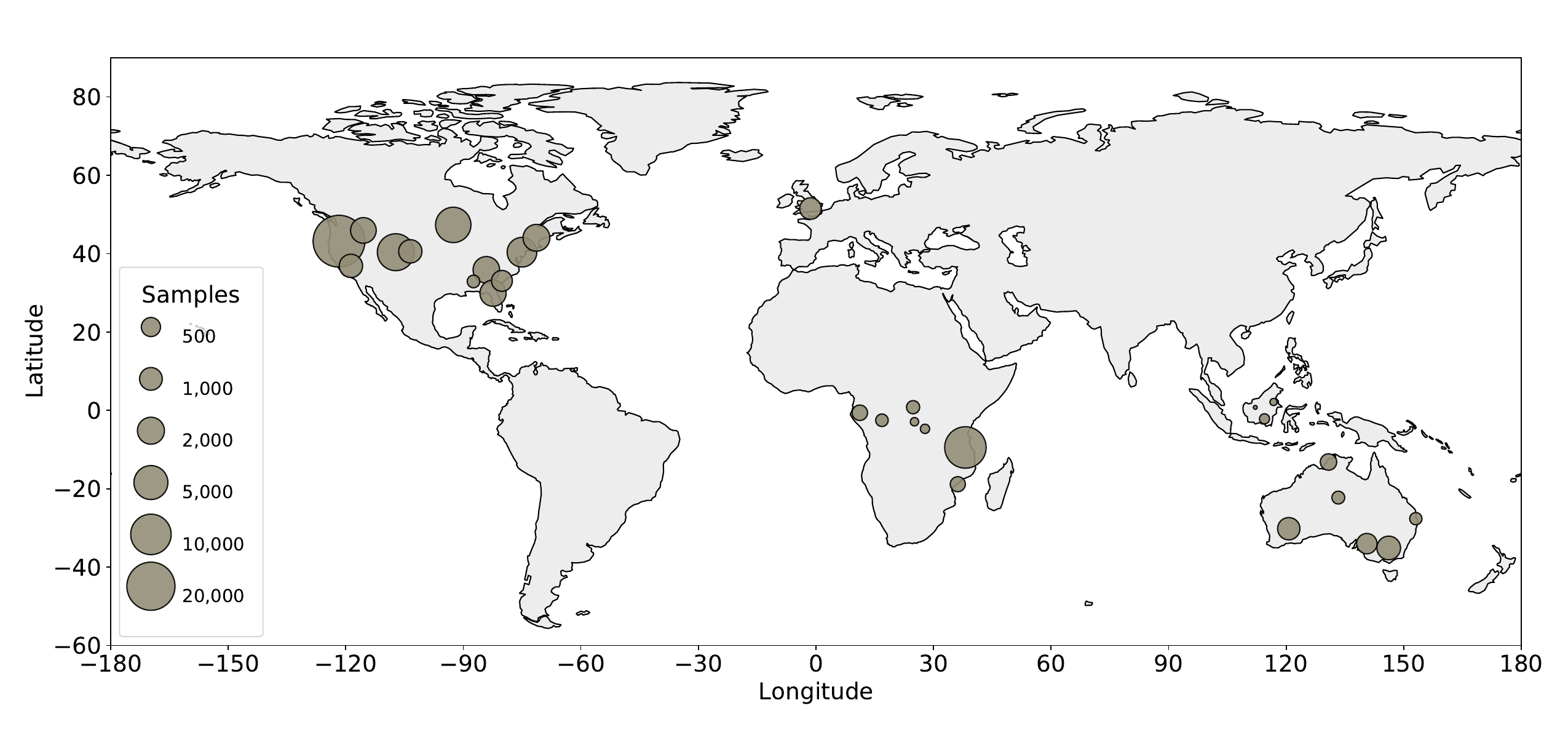}  }}%
    
    \subfloat[]{{\includegraphics[width=.98\textwidth]{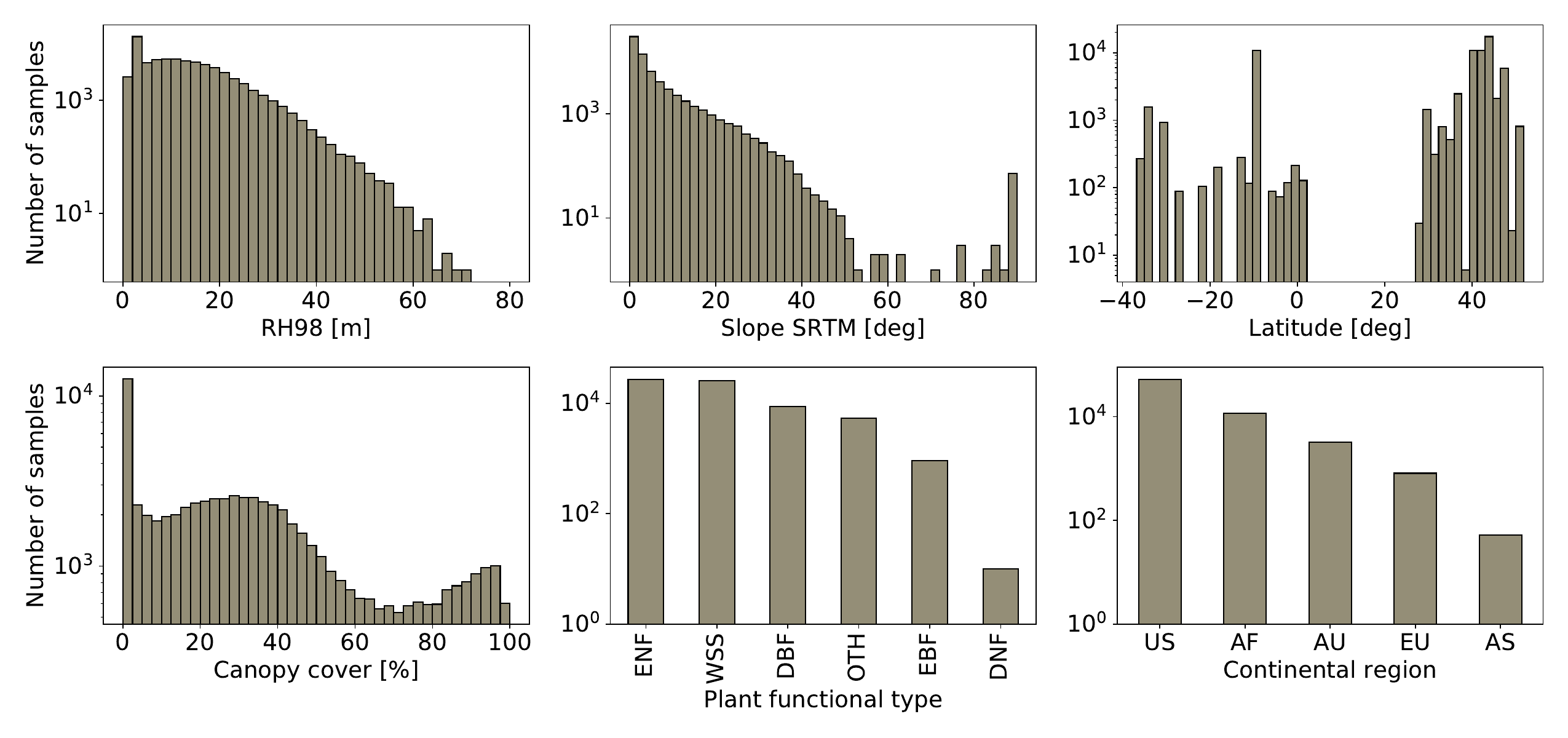} }}%
    \caption{Dataset of on-orbit GEDI waveforms with associated airborne LIDAR reference data (Pearson correlation \textgreater0.95). (a) Locations where GEDI waveforms are matched with airborne laser scanning (ALS) data. (b) Overall statistics of the dataset in terms of forest structure, slope, and continental region. The plant functional types are: ENF: Evergreen needleleaf forest, WSS: woodland, savanna, shrubland, DBF: Deciduous broadleaf forest, EBF: Evergreen broadleaf forest, DNF: Deciduous needleleaf forest, OTH: Other (water, urban, croplands)}
    \label{fig:gt_data}
\end{figure*}

To learn robust features that capture the global canopy height distribution and generalize to different biomes and forest types, the proposed CNN requires a representative dataset of corresponding waveforms.  
An extensive airborne LIDAR dataset has been collected by the GEDI mission~\citep{DUBAYAH2020100002} that spans five continental regions (Fig.~\ref{fig:gt_data}a) and covers a diverse set of forest structures in terms of canopy top height (RH98), canopy cover, and plant functional type (extracted from the MCD12Q1~\citep{sulla2018user}). The dataset also spans a broad range of topography \add{(flat to steep terrain)} between $-$36.8 and 51.8 degree latitude (Fig.~\ref{fig:gt_data}b). All ground slope data is derived from the Shuttle Radar Topography Mission (SRTM)~\citep{farr2007shuttle}.
\add{North America contributes more than half of the data. For South America and East Asia, there is no reference data available and Europe is scarcely represented in the current dataset.}

\add{To construct the training and validation dataset used in this work, GEDI L1B waveforms from the first seven months of the mission were matched with the ALS data.} 
Assigning reference data from ALS to the GEDI waveforms is complicated by geolocation errors in the first release of GEDI L1B data, which introduces noise into the reference \add{dataset}.
To correct the systematic geolocation error of the GEDI waveforms, sequences (blocks) of GEDI waveforms along individual ground tracks are correlated with simulated GEDI-like tracks of waveforms created from the intersecting airborne laser scanning (ALS) reference data~\citep{blair1999modeling,hancock2019gedi}. The process determines the horizontal and vertical offset that maximizes the correlation between on-orbit and simulated waveform sequences. 
For each ground track, a rigid block of successive GEDI waveforms is shifted to maximize the average Pearson correlation coefficient between the on-orbit waveforms and the simulated reference waveforms. After this transformation, the collocated GEDI and ALS reference data are saved along with the Pearson correlation coefficient of the individual waveforms.

Once we have filtered poor quality offsets (computed using \textless25 shots and with a correlation \textgreater0.9), we use the Pearson correlation coefficient between individual on-orbit and simulated waveforms to assess the quality of the match, and thus the quality of the assigned true canopy top height (RH98).
Remaining noise in the reference value assignment may exist due to simulation inaccuracies, temporal changes, and the random component of geolocation errors.
Empirically, we found that rejecting samples with \textless0.95 Pearson correlation yields largely clean reference data, which results in a total of 68,483 waveforms with a maximum RH98 of 70.1~m. 
For these data the systematic geolocation error for individual beams amount to 19.7$\pm$10.7~m (mean$\pm1\sigma$), but in the case of orbits with degraded geolocation flags offsets may reach up to 60~m.
When applying a higher Pearson threshold, fewer samples (with presumably higher quality) remain in the dataset, e.g., an overly strict threshold of 0.99 would \add{discard a substantial amount of samples (only 19,186 samples retain in the dataset) leading to a less representative dataset in which no samples above 50~m RH98 would be retained.}

\section{Method}

\subsection{Waveform preprocessing}
Several preprocessing steps are applied to the raw L1B waveforms to harmonize the input features. After subtracting the mean noise level of the GEDI L1B product, we normalize the total energy of each waveform to one. Additionally, we set all waveforms to \add{the fixed maximum} length of $1420$ vertical bins \add{by padding} shorter waveforms with zeros at the end of the vector.
A fixed waveform length allows the CNN to implicitly learn the absolute scale of canopy heights, as the interval between consecutive returns is fixed to 15~cm.
We \add{normalize} the distribution of the input features (waveform amplitudes) as well as the targets (canopy heights) to standard normal distributions, a common preprocessing step for neural networks~\citep{lecun2012efficient}. Since the learnable network parameters are randomly initialized in such a way that the distribution of the input data is preserved in the output, that preprocessing leads to faster convergence during optimization. The means and standard deviations of the training data are stored and used to harmonize the test data in the same way.

\subsection{Convolutional neural network architecture}
Our method uses a deep CNN at its core to regress canopy top height. We adapt the ResNet architecture~\citep{he2016deep} to our needs by, first, replacing all two-dimensional network layers with one-dimensional convolutional, pooling and normalization layers. Second, we found empirically that a network architecture with more regular down-sampling than the original ResNet performs better. More precisely, our \textit{SimpleResNet} consists of eight residual blocks (Fig.~\ref{fig:CNN_illustration} in the appendix). Each block consists of two consecutive convolutional layers with kernel size $3$, each followed by batch normalization and a \textit{rectified linear unit} (ReLU) activation function. All blocks include a \textit{skip connection} that adds the input of the block to its final activation map. Such a bypass not only alleviates the vanishing gradient problem of deep neural networks, but simplifies the task of a block to learning a (usually smaller) residual correction to its input.
After each residual block, a max-pooling layer reduces the extent of the intermediate feature representations (i.e., \textit{activation maps}) by a factor of two. This reduces memory consumption and speeds up the computation of the network output. Furthermore, down-sampling gradually increases the \textit{receptive field} of the learned features, so that deeper features represent larger data segments of the input waveform, encoding more contextual evidence. The last block is followed by an adaptive average pooling layer that averages each final feature map to a scalar value, resulting in a 1-dimensional feature vector. A dropout layer~\citep{srivastava2014dropout} with drop ratio of 0.5 is used for regularization, before the final, fully connected layer creates the model outputs.

\subsection{Predictive uncertainty estimation}
Uncertainty in deep neural networks originates from two main sources. While the \textit{aleatoric uncertainty} describes the stochasticity in the data, the \textit{epistemic uncertainty} captures uncertainty of the model, i.e., the ambiguity of its individual parameters. Aleatoric uncertainty is represented with a probability distribution over the model output, describing the conditional probability of different target values given the input, $p(y|x)$. Epistemic uncertainty is represented as a probability distribution over the model parameters~\citep{gast2018lightweight}. The epistemic uncertainty arises from a lack of knowledge, in the case of a supervised learning task due to limited training data. Therefore, epistemic uncertainty can be reduced by collecting additional data that adds information not present in the current training set. In contrast, aleatoric uncertainty is inherent in the problem and cannot be reduced. We model it by approximating $p(y|x)$ with a Gaussian probability density function. To estimate the epistemic uncertainty, we use an ensemble of ten CNNs~\citep{lakshminarayanan2017simple}. \add{Note that an ensemble of five models may already be enough to estimate the epistemic uncertainty~\citep{ovadia2019can}.}

In general, the goal is to approximate the true conditional probability distribution $p(y|x, \theta)$ of the target value $y$ (e.g., the ALS canopy height), given an observed input waveform $x$. To that end we fit the parameters $\theta$ of a neural network $f_\theta$, which outputs a prediction  $f_\theta(x)$. The most common approach is to train the network to predict a single point-estimate $\hat{y}=f_{\theta}(x)$, by minimizing the mean squared error (MSE) between the true values $y$ and the predicted values $\hat{y}$ ~\citep{Goodfellow-et-al-2016}. 
Formally, optimizing the MSE is equivalent to parameterizing the conditional distribution as a Gaussian normal distribution $p(y|x) = \mathcal{N}(\hat{y}(x), \sigma^2)$, with the mean $\hat{y}(x)$ as a function of the input $x$ and a constant noise term $\sigma^2$~\citep{Goodfellow-et-al-2016}. We relax this constraint of constant noise and allow more flexibility, by also estimating the noise from the input. Instead of a point-estimate $\hat{y}$, our network has two outputs $[\hat{\mu}, \hat{\sigma}^2]=f_{\theta}(x)$ to approximate the true conditional distribution with a Gaussian $p(y|x) = \mathcal{N}(\hat{\mu}(x), \hat{\sigma}^2(x))$, with both mean~$\hat{\mu}$ and variance~$\hat{\sigma}^2$ as functions of the input $x$~\citep{gast2018lightweight}.
We thus minimize the Gaussian negative log likelihood (NLL) as follows:
\begin{equation}
    \centering
    \mathcal{L}_{NLL} = \frac{1}{N} \sum_{i=1}^{N} \frac{ \left ( \hat{\mu}(x_i) -y_i   \right )^2}{2 \hat{\sigma}^2(x_i) } + \frac{1}{2}\log \hat{\sigma}^2(x_i).
    \label{eq:gaussian_NLL}
\end{equation}
Within Bayesian estimation, this is also known as measuring the \textit{heteroscedastic} aleatoric uncertainty~\citep{kendall2017uncertainties}. In practice, our network outputs the two parameters $\mu$ and $s$, with $s = \log \sigma^2$.
Converting the second output to the variance $\sigma^2=\exp (s)$ guarantees that the estimated variance is positive~\citep{kendall2017uncertainties}. For numerical stability, a small number $\epsilon =10^{-8}$ is added to the estimated variance, to avoid division by zero. 

\begin{figure*}[]
    \centering
    \includegraphics[width=\textwidth]{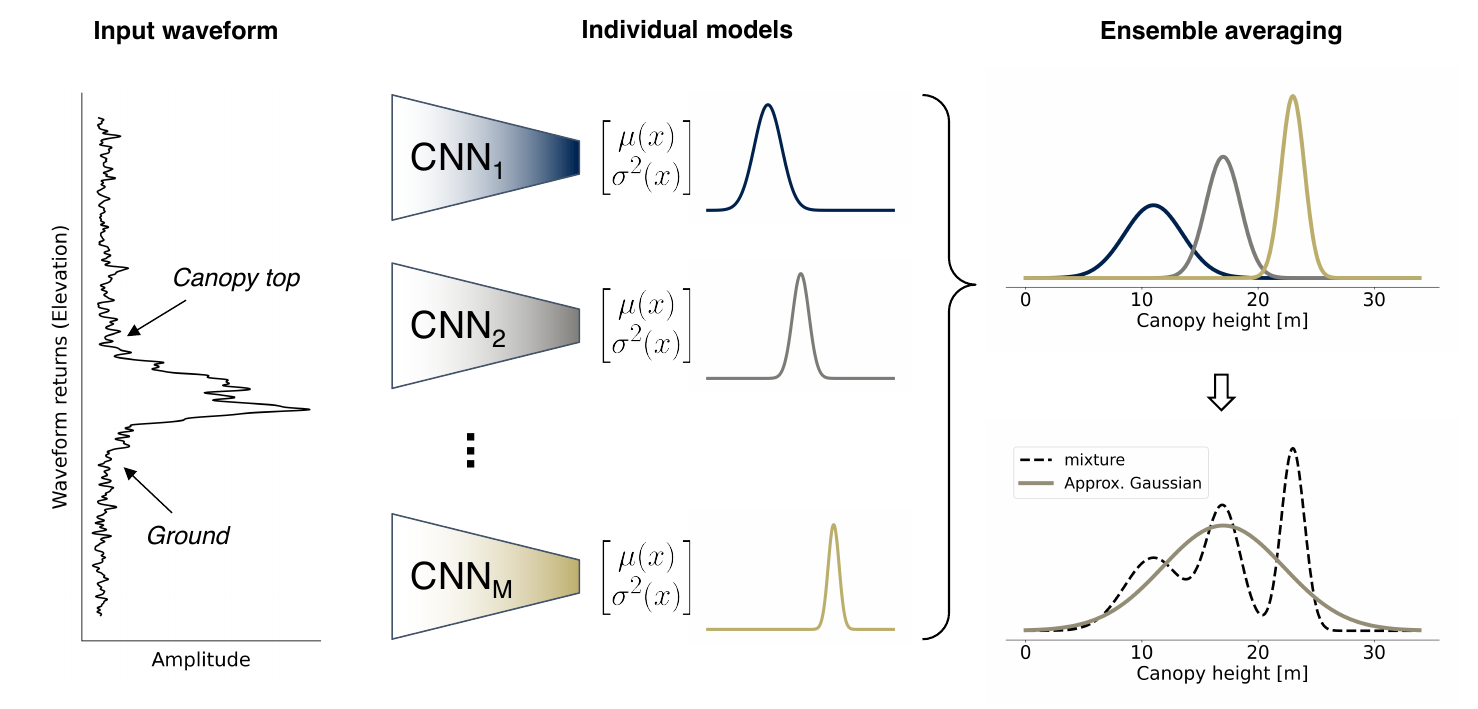}
    \caption{Illustration of the model ensemble. A L1B GEDI waveform is fed into multiple CNNs that have identical architecture, but were trained independently. Each CNN estimates a mean and a variance of the conditional distribution of canopy height. Averaging the individual outputs yields a mixture distribution, whose mean and variance constitute the final estimate.}
    \label{fig:model_ensemble}
\end{figure*}

We model the epistemic uncertainty with an ensemble of deterministic CNN models (illustrated in Fig.~\ref{fig:model_ensemble}), which is computationally more efficient than Bayesian neural networks, and tends to perform better in practice~\citep{lakshminarayanan2017simple,ovadia2019can}. 
\add{Deep ensembles provide an effective approach to approximate Bayesian marginalization, i.e. marginalizing out parameters instead of optimizing towards a single “best” parameter~\citep{gustafsson2020evaluating,wilson2020case,NEURIPS2020_322f6246}.}\footnote{\addtwo{There appears to be disagreement about whether deep ensembles belong to the Bayesian family of models. While \cite{lakshminarayanan2017simple} have described deep ensembles as an alternative to Bayesian neural networks that model the distribution over weights, \cite{NEURIPS2020_322f6246} have shown that deep ensembles provide an effective mechanism to approximate the Bayesian predictive distribution.}}
The basis of this strategy is to train $M$ different networks and regard them as random samples from the distribution of models. We do this by starting the training with a different random initialization for each network, along with the randomness due to learning with stochastic gradient descent. At prediction time we run the input waveform through each network and compute the variance of the $M$ outputs. Intuitively, each individual network should generalize about equally well, as long as a test sample lies within the training data distribution. In contrast, test samples that are out-of-distribution should cause arbitrary errors that differ between the $m$ networks, and therefore have high variance. While reliably detecting \add{out-of-distribution (OOD)} samples is an unsolved problem and and active direction of research, deep ensembles at present perform best in practice.

Averaging outputs of the individual CNNs corresponds to a mixture of Gaussians with equal weight, as illustrated in Fig.~\ref{fig:model_ensemble}. The final prediction of the model ensemble is reported as the mean $\hat{y}$ (Eq.~\ref{eq:mean_prediction}) and the variance $Var(\hat{y})$ (Eq.~\ref{eq:predictive_var}) of this mixture distribution:
\begin{equation}
     \hat{y} = \frac{1}{M} \sum_{m=1}^{M} \hat{\mu}_{m},
    \label{eq:mean_prediction}
\end{equation}
\begin{equation}
     Var(\hat{y}) = \frac{1}{M} \sum_{m=1}^{M} \hat{\mu}_{m}^2 - \left ( \frac{1}{M} \sum_{m=1}^{M} \hat{\mu}_{m}  \right )^2 + \frac{1}{M} \sum_{m=1}^{M} \hat{\sigma}_{m}^2,
    \label{eq:predictive_var}
\end{equation}
with input $x$ and output $[\hat{\mu}_m, \hat{\sigma}_m]$ of model $m=[M]$. 
Under our model, the variance of this Gaussian mixture distribution is the \textit{predictive uncertainty}, composed of the epistemic uncertainty (first two terms in Eq.~\ref{eq:predictive_var}) and the aleatoric uncertainty (last term in Eq.~\ref{eq:predictive_var}).

\subsection{Model training}
We split the dataset into non-overlapping training and test sets. Within the training set, a random subset of 10\% is set aside as validation set to monitor the optimization process. Thus, CNN parameters are fitted to 90\% of the training set. The validation set is not used in the optimization, but serves to monitor the loss (cumulative prediction error) on unseen data. For each ensemble member, the network parameters with the lowest validation loss are used in the final model, which is then applied to the held-out test set to evaluate model performance.
For each individual CNN in the ensemble the learnable parameters (weights) are initialized randomly, and trained by minimizing the Gaussian negative log-likelihood as a loss function (Eq.~\ref{eq:gaussian_NLL}).
That minimization is carried out with ADAM~\citep{KingmaB14}, an adaptive version of stochastic gradient descent (SGD). In each iteration of the optimizer, the loss is computed on a random batch of 64 data samples. Based on that loss, partial derivatives (gradients) with respect to all model parameters are determined by back-propagation, i.e., calculation of the numerical gradients with the chain rule.
Model parameters are updated in small steps along the negative gradient direction to minimize the loss. The step size is controlled by a hyper-parameter called the \textit{learning rate}, which we initialize to 0.0001. ADAM then automatically adapts the learning rate after every iteration. Each model is trained for 200 epochs, with one epoch corresponding to the number of iterations (respectively, random batches) needed to go once through the complete training set.
Since shifting the waveform up and down should not affect the canopy top height, we apply random shifts to augment the training data. This can be seen as simulating canopies with the same LIDAR response at different elevations. By drawing shifts uniformly at random from $\pm$20\% of the waveform length, we ensure that the network predictions are invariant against such shifts. 

\subsection{Random cross-validation}
We perform random 10-fold cross-validation to measure whether the model performance is invariant against different train/test splits. The complete dataset of waveforms with ground reference is split randomly into 10 mutually exclusive subsets. Each subset serves once as the test set, while the remaining nine subsets are used to train the ten CNN models from scratch, i.e. the full ensemble is retrained ten times and evaluated on the corresponding, held-out test fold.
Overall performance metrics are calculated over all ten cross-validation folds, such that each waveform was in the test set exactly once.
Additionally, we measure the variability of the error metrics between the ten test sets. 

\subsection{Geographical cross-validation}
Geographic transferability of the model is tested by splitting into geographical regions. We reserve all available samples from one geographic region for testing and train the models from scratch on the other regions, such that no sample from the test region has been seen during training.
With this procedure one can assess whether the model over-fits to the training locations, or whether it learns generic features that also support prediction in unseen locations. 

\subsection{Evaluation metrics}
As main error measure to quantify the deviations between predicted heights and reference heights, we report the root mean square error (RMSE). Moreover, we measure the bias of the prediction with the mean error (ME):
\begin{equation}
    \centering
    RMSE = \sqrt{ \frac{1}{N} \sum_{i=1}^{N} \left ( \hat{y}_i - y_i   \right )^2 },
    \label{eq:RMSE}
\end{equation}
\begin{equation}
    \centering
    ME = \frac{1}{N} \sum_{i=1}^{N} \left ( \hat{y}_i - y_i \right ),
    \label{eq:ME}
\end{equation}
where $\hat{y}_i$ denote the prediction of the model ensemble and $y_i$ the reference value at sample $i$.
Under this definition, a positive ME means that model predictions are higher than the true values according to reference data. For completeness and to allow comparisons with other works, we additionally report the mean absolute error (MAE):
\begin{equation}
    \centering
    MAE = \frac{1}{N} \sum_{i=1}^{N} \left | \hat{y}_i - y_i \right |.
    \label{eq:MAE}
\end{equation}
To study relative deviations (height error in percent of the reference height) we use the mean absolute percentage error (MAPE):
\begin{equation}
    \centering
    MAPE = \frac{100}{N} \sum_{i=1}^{N} \left | \frac{\hat{y}_i - y_i}{y_i} \right |,
    \label{eq:MAPE}
\end{equation}
which normalizes the residuals by the "true" height before computing their absolute differences. 

\subsection{Evaluation of the predictive uncertainty calibration}
\add{The predictive uncertainty can be used to rank the predictions of individual samples. A standard tool for the evaluation of ranked retrieval results are precision-recall curves \citep{schutze2008introduction}.}
Beside precision-recall curves, calibration plots are a common strategy for analyzing the estimated predictive uncertainty~\citep{guo2017calibration}. 
The estimated predictive standard deviation is plotted against the empirical error by marginalizing over subsets of samples. The closer the resulting line in the diagram follows the diagonal, the better is the calibration of the uncertainties. 
Specifically, for regression tasks with continuous output values (like our canopy height) we expect the predictive standard deviation to follow the RMSE~\citep{laves2020well}. 
In order to get robust, statistically meaningful empirical errors, we must group the test samples. We do this by binning the samples into 1~m intervals based on their predictive standard deviation. For all bins with \textgreater200 samples the average predictive STD is plotted against the RMSE (Fig.~\ref{fig:uncertainty_vs_error}b).

\section{Results}
\subsection{Canopy top height regression}

\begin{figure*}[]
    \centering
    \subfloat[]{{\includegraphics[width=0.45\textwidth]{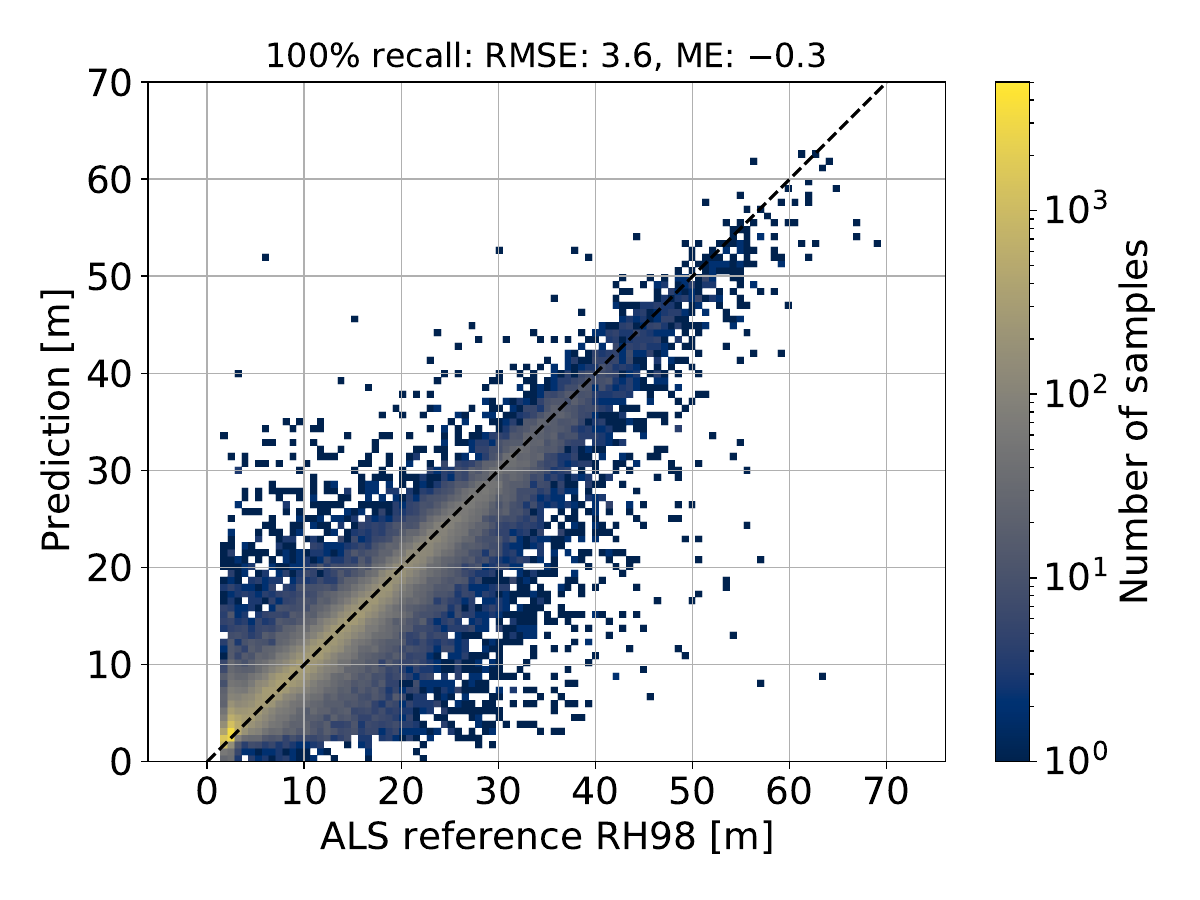} }}%
     \subfloat[]{{\includegraphics[width=0.45\textwidth]{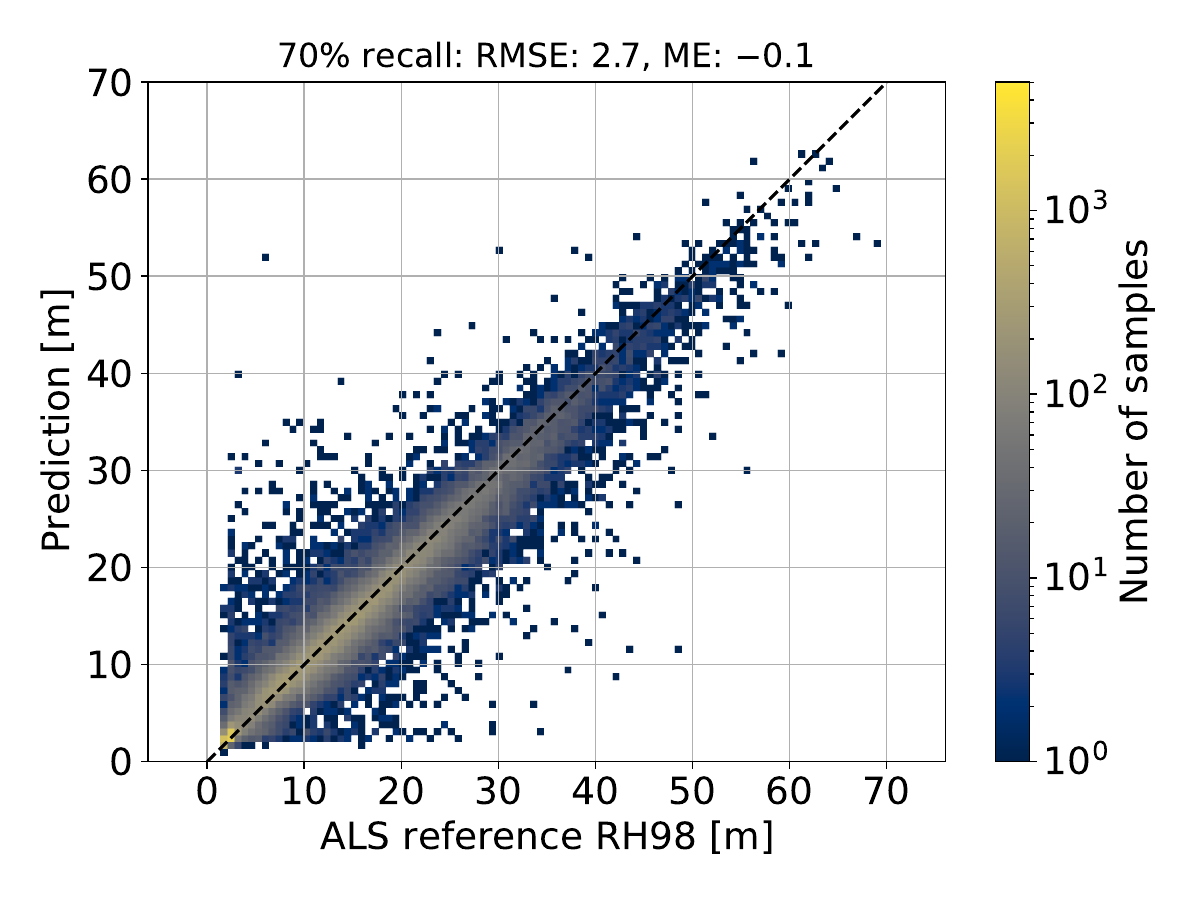} }}%
     
    \caption{Regression of canopy top height (RH98) from L1B GEDI full waveforms. ALS reference data vs. CNN predictions for (a) all test samples. (b) the 70\% most certain test samples, filtered based on the predictive standard deviation with adaptive thresholds (depending on the predicted canopy height).}
    \label{fig:regression_rh98}
\end{figure*}

\begin{figure*}[]
    \centering
    \includegraphics[width=\textwidth]{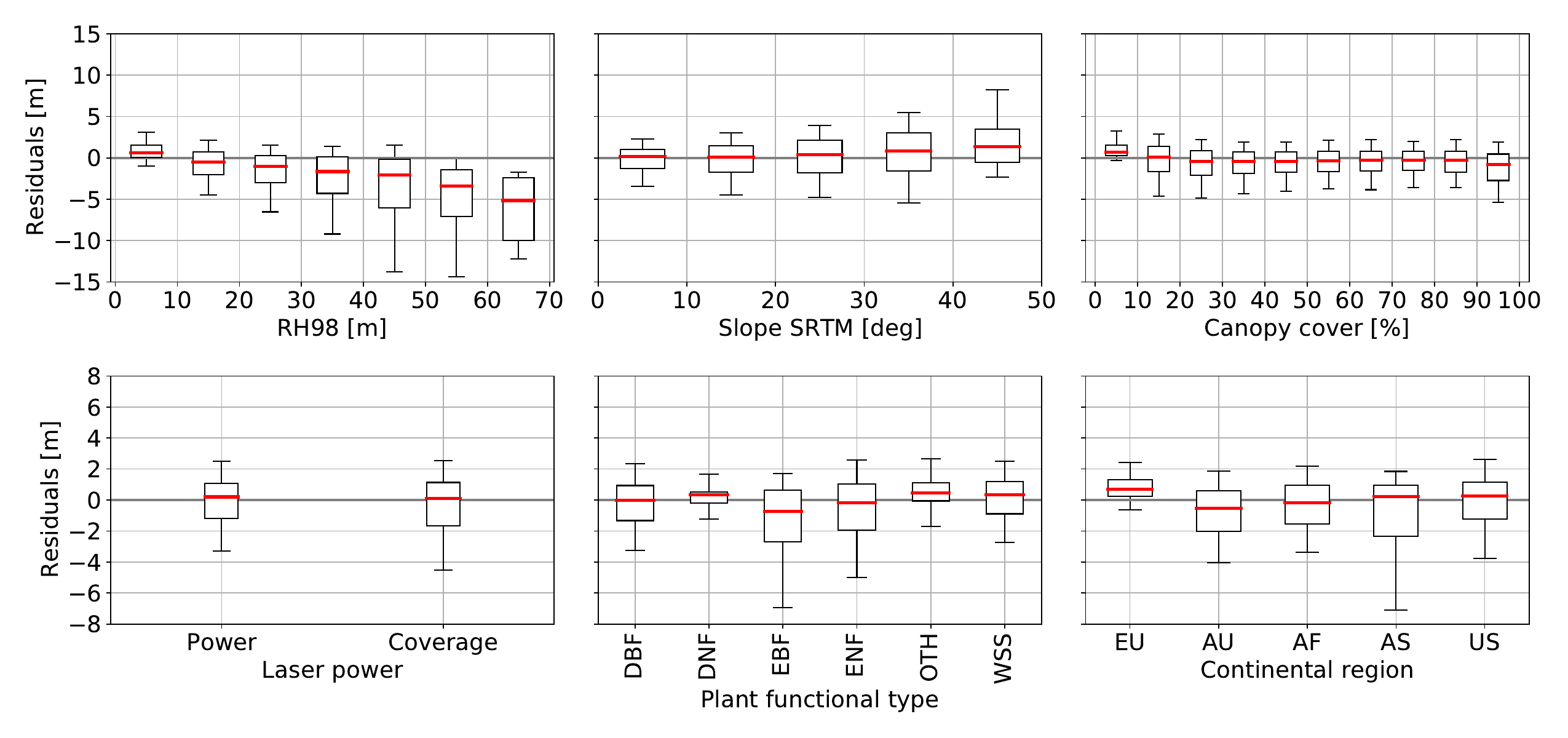}
    \caption{Analysis of residuals for canopy top height (RH98) regression (results of 10-fold random cross-validation). 
    Positive residuals denote that the predictions are higher than the reference values.
    The boxplots show the median, the quartiles, and the 10\textsuperscript{th} and 90\textsuperscript{th} percentiles. \add{The plant functional types are: 
    DBF: Deciduous broadleaf forest, 
    DNF: Deciduous needleleaf forest, 
    EBF: Evergreen broadleaf forest, 
    ENF: Evergreen needleleaf forest, 
    OTH: Other (water, urban, croplands),
    WSS: woodland, savanna, shrubland.} 
    The continental regions are: EU: Europe, AU: Australia, AF: Africa, }
    \label{fig:residual_analysis}
\end{figure*}

Random cross-validation on this dataset yields a root mean square error (RMSE) of 3.6~m and a mean error (ME) of $-$0.3~m, indicating a slight underestimation bias compared to the ALS reference (Fig.~\ref{fig:regression_rh98}a). 
To investigate whether the performance is affected by specific data splits, the variation of the error metrics across the ten random folds was analyzed  (Tab.~\ref{tab:geograph_generalization}). Both the RMSE and the ME have 0.1~m standard deviations across the ten folds, i.e., the performance is invariant to the random data splits. The mean absolute percentage error (MAPE) accounts for 26.5~\% averaged over all test samples, where samples with low reference values inherently have high relative errors.
A detailed residual analysis with respect to forest structure and topography suggests that the canopy height estimates are robust against diverse characteristics (Fig.~\ref{fig:residual_analysis}). Negative residuals denote an underestimation of the canopy top height compared to the ALS reference.
We observe a slight positive bias for samples with RH98\textless10~m and a systematic negative bias (underestimation) for higher canopies \add{(RH98\textgreater10~m)}. Yet the median of the residuals is \textless10~\% ($\approx$5~m) for canopies \textgreater60~m. A slightly positive bias appears with slope\textgreater30 degrees. 
Both systematic effects seem to be related to underrepresented samples in our dataset (Fig.~\ref{fig:gt_data}b).
The predictions are less affected by canopy cover or plant functional type. Additionally, no considerable biases are observed regarding latitude and continental region \add{(Fig.~\ref{fig:residual_analysis})}. 
For both the power and coverage lasers, the residuals have a median close to zero (Fig.~\ref{fig:residual_analysis}). While the overall performance is similar for different laser powers, further investigations reveal that laser power affects the performance for high canopies (Fig.~\ref{fig:boxplot_coverage_high_canopy}a in the appendix). Specifically, the samples with reference RH98\textgreater50~m have a stronger underestimation bias with the coverage laser (ME of $-$12.1~m) compared to the power laser (ME of $-$5.8~m).
Finally, the overall CNN performance (over all heights) is not affected by daylight background noise and appears to be independent of beam sensitivity (Fig.~\ref{fig:boxplot_solar_and_sensitivity} in the appendix).

\subsection{Geographic generalization}
Deep neural networks have large model capacity, i.e., enough trainable parameters to adapt to details of the training data distribution. It is thus important to verify that our model does not over-fit to particular training regions. Therefore we conducted experiments with multiple geographic train/test splits to investigate geographical generalization (also called geographic transferability~\citep{DUBAYAH2020100002}). In each experiment, all samples from a particular test region are used for evaluation and the model is trained from scratch without seeing any sample from that test region. 
We present the geographical generalization results for Europe, Australia, Africa, and the tropics (between 23.5 degrees north and south) in Tab.~\ref{tab:geograph_generalization}, the corresponding scatter plots are included in the appendix (Fig.~\ref{fig:geographic_generalization}).

None of those experiments resulted in a higher RMSE. 
The bias is slightly stronger than with random cross-validation without geographical generalization (Random CV) with a mean error (ME) of $-$0.3~m. Canopy height is overestimated in Europe with a ME of 0.7~m, but underestimated in Australia, Africa and the tropics with a ME of $-$0.9~m. 
The relative error (MAPE) grows from 25.6\% for the random cross-validation to 45.9\% in Europe and to 32.1\% in Australia. For both Africa and the tropics, it slightly decreases to 24.2\%. The reason for this is that in Europe and Australia the canopy heights are largely below 30~m, and with decreasing height the computation of the relative error is increasingly dominated by the constant (not height-dependent) part of the absolute error.
(see Fig.~\ref{fig:geographic_generalization} in the appendix).
\add{When using all African regions as the test set, the model has seen some data from tropical Asia, compared to the last experiment that holds out all available data from tropical regions as the test set. However, we do not observe any notable difference in performance.} Overall, our experiments suggest that the proposed model generalizes reasonably well to geographical regions not seen during training and is not over-fitted to specific (training) regions.

\begin{table*}[]
    \centering
    \begin{tabular}{lrrrr}
    \toprule
    {} &  RMSE [m] &  MAE [m] &  ME [m] &  MAPE [\%] \\
    \midrule
    Random CV &       3.6 &      2.1 &    -0.3 &       25.6 \\
     &     (0.1) &    (0.0) &   (0.1) &      (0.9) \\
    Europe    &       2.5 &      1.5 &     0.7 &       45.9 \\
    Australia &       3.2 &      2.1 &    -0.9 &       32.1 \\
    Africa    &       3.1 &      2.0 &    -0.9 &       24.2 \\
    Tropics   &       3.1 &      2.1 &    -0.9 &       24.2 \\
    \bottomrule
    \end{tabular}
    \caption{Geographic generalization: The first row shows random cross-validation results as baseline, with the standard deviation over ten cross-validation folds in parentheses. The remaining rows correspond to geographic generalization experiments, where the listed continental region was the test data and was held out during training.}
    \label{tab:geograph_generalization}
\end{table*}

\subsection{Estimation of the predictive uncertainty}
\begin{figure*}[]%
    \centering
    \subfloat[]{{\includegraphics[width=0.45\textwidth]{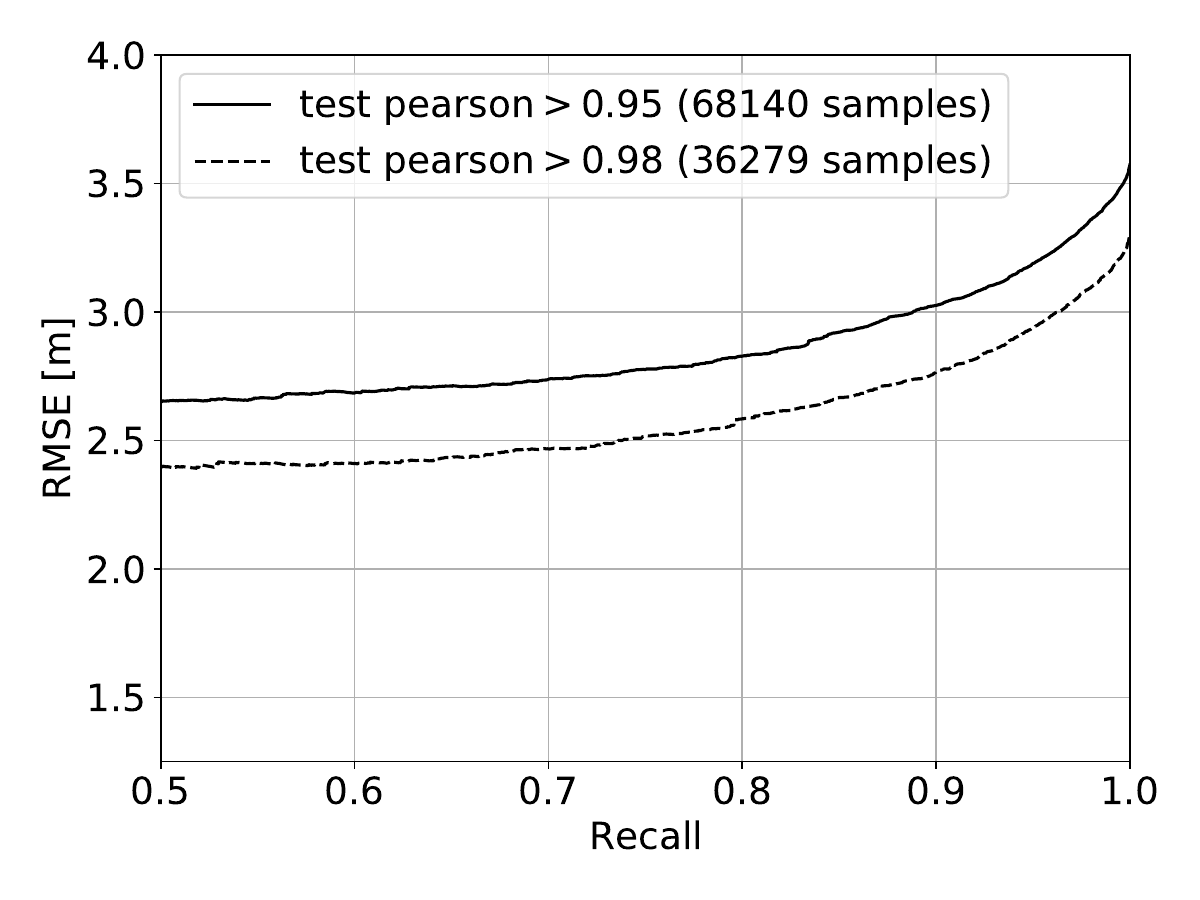} }}%
    \subfloat[]{{\includegraphics[width=0.45\textwidth]{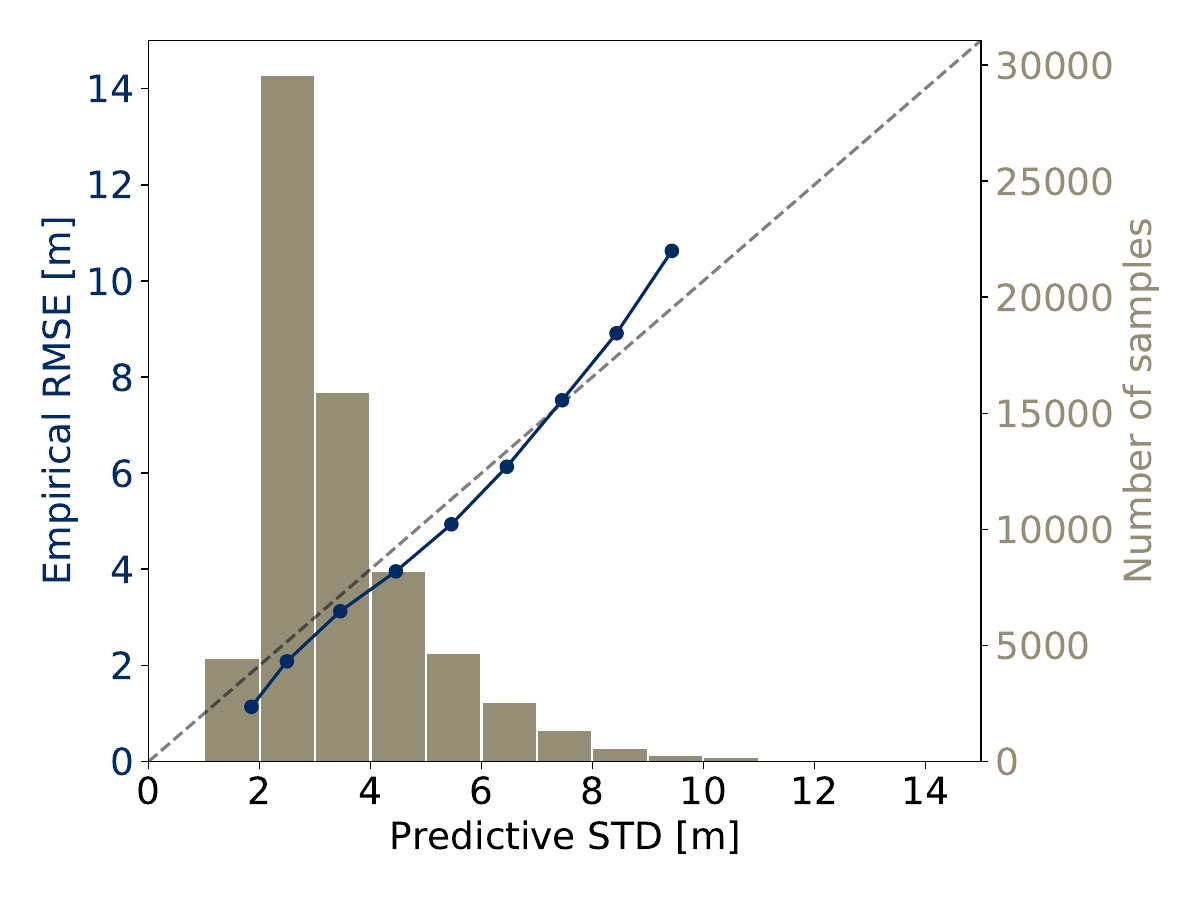} }}%
    
    \caption{Evaluation of the predictive uncertainty with respect to the empirical error. (a) Precision-recall plot filtering the test data based on the predictive uncertainty. \add{Recall describes the fraction of samples retained.} (b) Calibration plot for the predictive uncertainty of RH98 showing the estimated standard deviation (STD) vs.\ the empirical RMSE for subsets grouped by the predictive STD.}%
    \label{fig:uncertainty_vs_error}%
\end{figure*}

The estimated predictive uncertainty should, in expectation, correlate with the empirical error~\citep{guo2017calibration}. We employ two different schemes to evaluate the quality of the estimated uncertainty (Fig.~\ref{fig:uncertainty_vs_error}).
One intuitive way to analyze whether the estimated uncertainty is meaningful is to sort the test samples by their predictive uncertainty. Filtering of the test samples with increasingly strict thresholds for the highest permissible uncertainty should then progressively reduce the overall error (Fig.~\ref{fig:uncertainty_vs_error}a). We observe that the RMSE initially goes down quickly as recall decreases, then flattens and becomes stable against recall.
First, we evaluate this filtering over all test samples with a Pearson correlation coefficient \textgreater0.95 \add{(resulting from the waveform matching)}. Removing the most uncertain 10\% of the predictions reduces the RMSE from 3.6~m (100\% recall) to 3.0~m (90\% recall). For example, at the 70\% recall, the RMSE reaches 2.7~m. 
Filtering based on the estimated uncertainty effectively reduces the overall error, which shows that the model uncertainty is predictive of empirical error.
Secondly, we apply a stricter Pearson correlation threshold of 0.98 to the test data, keeping only those 53\% of the test samples for which the reference data has (presumably) higher quality. We observe that in that setting the RMSE decreases by $\approx$0.25~m over the entire range of recall, i.e., at 70\% recall the RMSE falls below 2.5~m. 
This behavior indicates remaining inaccuracies in the reference data due to simulator error, GEDI/ALS temporal mismatch, imprecise geolocation, and related factors that are not resolved by the waveform matching. This is directly reflected in the quantitative error metrics, even though its causes are not deficiencies of the CNN model, but of the test reference data.

The second evaluation looks directly at the calibration of the uncertainty estimates. Those estimates are well calibrated if they correlate linearly with the empirical error (RMSE). Our results illustrate that the standard deviations predicted by the \rem{Bayesian }CNN are well calibrated (Fig.~\ref{fig:uncertainty_vs_error}b).
For instance, samples for which the predicted standard deviation is 3.5~m indeed have an empirical RMSE \add{of $\approx3.5~m$ with respect to reference data}.
\add{Comparing the calibration plots of the aleatoric and epistemic uncertainty on its own, shows that the aleatoric uncertainty is the dominating source in the random cross-validation experiment (Fig.~\ref{fig:calibration_plots} in the appendix). Nevertheless, the epistemic uncertainty contributes a non-negligible part and may become more prominent for under-represented cases.}

\begin{figure*}[]%
    \centering
    \subfloat[]{{\includegraphics[width=0.45\textwidth]{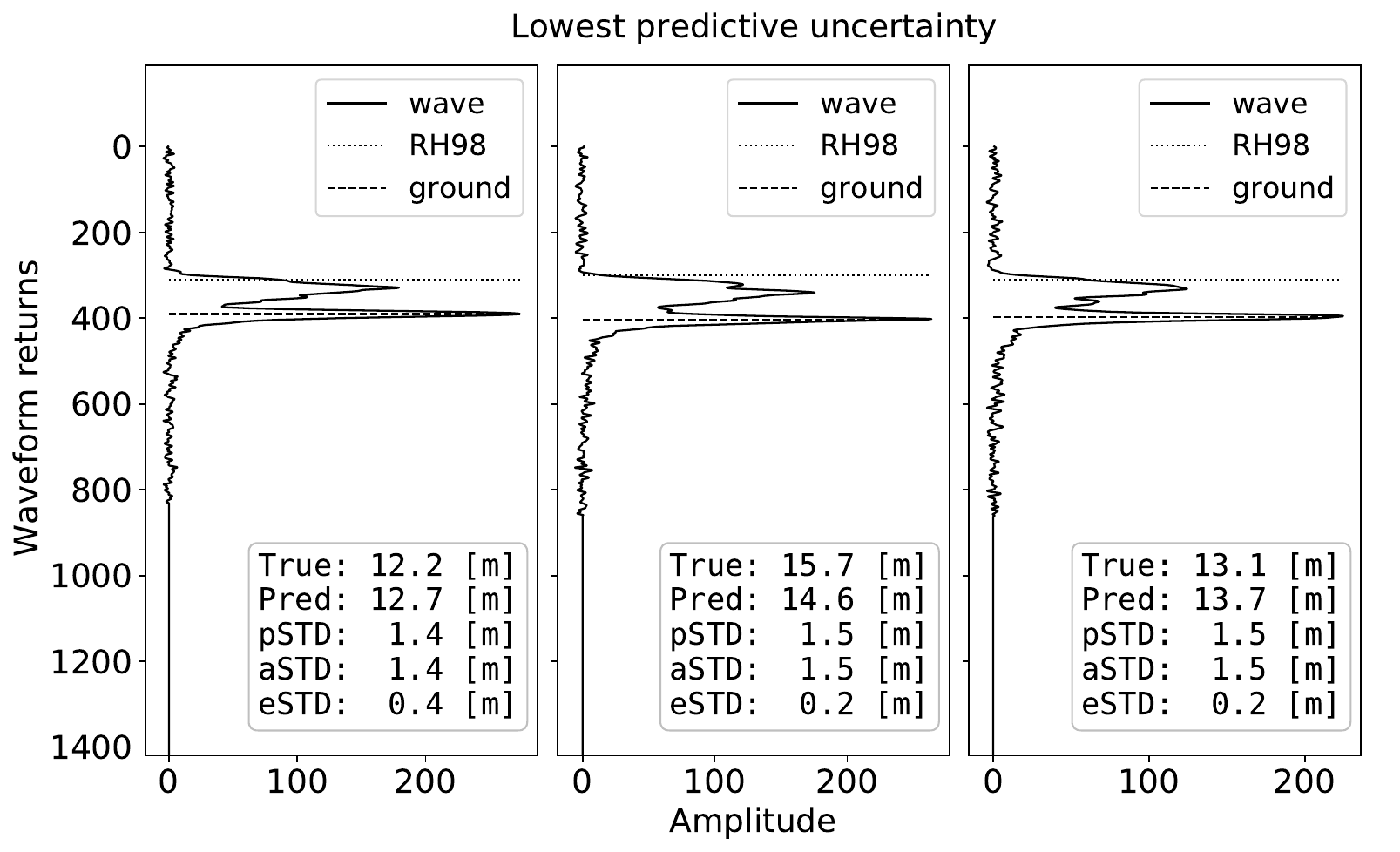} }}%
    \subfloat[]{{\includegraphics[width=0.45\textwidth]{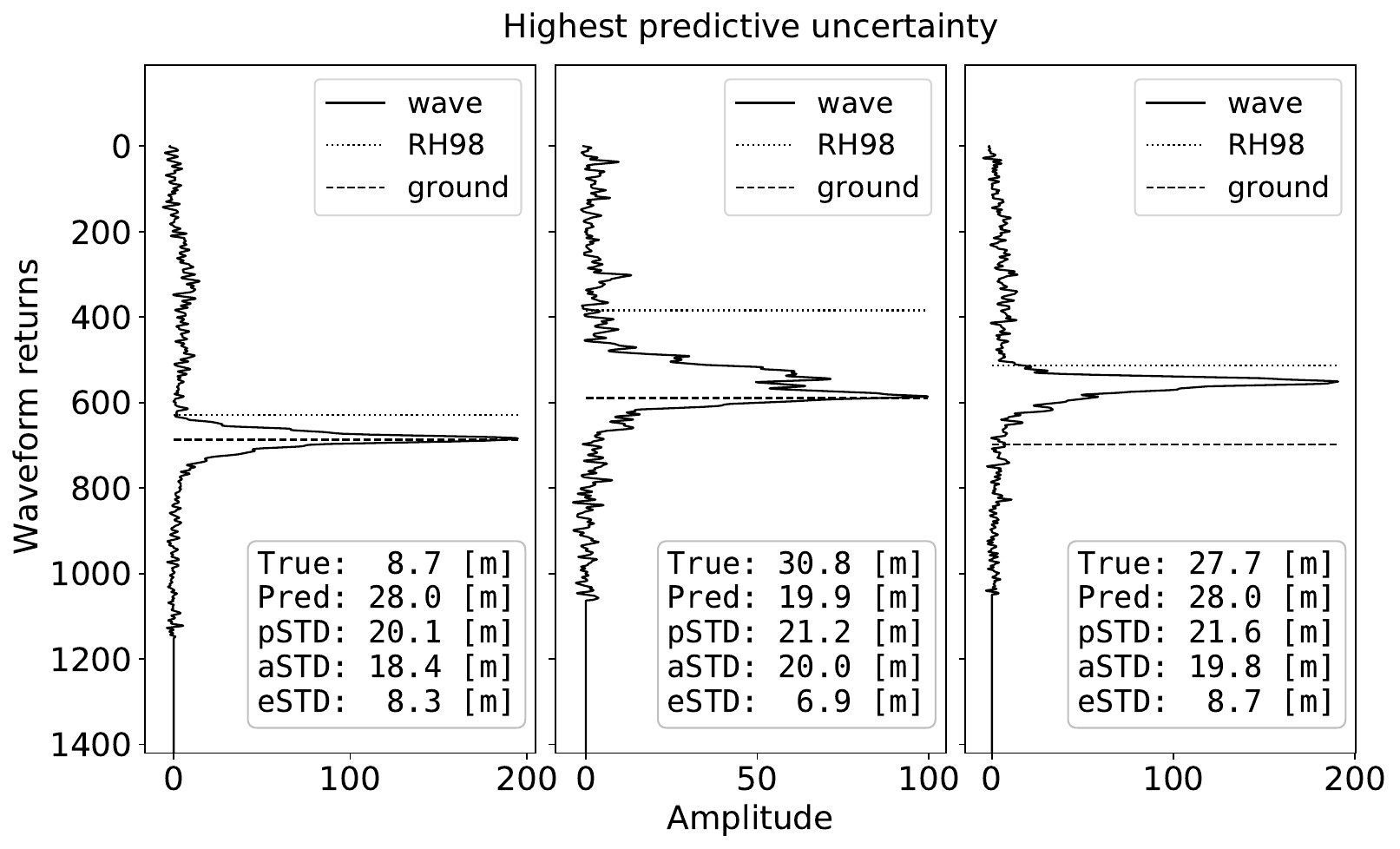} }}%
    \caption{Qualitative examples for the predictive uncertainty. (a) The waveforms with the lowest uncertainty. (b) The waveforms with the highest uncertainty. The RH98 and ground elevation correspond to the reference ALS data. The textbox lists: ALS reference canopy height (True), predicted canopy height (Pred), predictive standard deviation (pSTD), aleatoric standard deviation (aSTD), and epistemic standard deviation (eSTD).}%
    \label{fig:qualitative_uncertainty}%
\end{figure*}

Qualitatively, we observe that the CNN assigns lowest predictive uncertainty to waveforms with unambiguous signal for the top of the canopy (signal start) and the ground (last mode). In contrast, samples with particularly high predictive standard deviation correspond to waveforms where the canopy top and ground, as features, are less distinct. (Fig.~\ref{fig:qualitative_uncertainty}). 
This may occur for a variety of reasons, e.g. waveform noise, such as from background illumination, low-lying dense canopy that may obscure the ground, or canopy shape, such as in conifers, where the amount of reflective leaf material near the top is limited. 
While the CNN is not immune to \add{those effects that also hinder traditional waveform processing, our results illustrate the general advantage of a \rem{Bayesian}\addtwo{probabilistic} approach to make reliable predictions, i.e., the model knows in what cases it may be inaccurate.}

\subsection{Filtering canopy height estimates based on the predictive uncertainty}
A key advantage of the \rem{Bayesian}\addtwo{probabilistic} approach is that predictions can be filtered according to the estimated predictive uncertainty, so as to reduce the overall error at the cost of recall. The question is how to best design the corresponding filter. One important aspect for several applications (e.g., estimation of biomass) is that the full range of existing canopy heights should be preserved.
We find that both filtering based on the absolute and the relative standard deviation (i.e., the coefficient of variation) have limitations.
Such naive filters reduce the overall error, but effective thresholds on the absolute uncertainty remove almost \emph{all} high canopies. In contrast, thresholding the relative uncertainty removes almost all low canopies (see Fig.~\ref{fig:filtering_abs_rel} in the appendix).

To preserve the full height range, but still reduce the  expected error, we apply an adaptive threshold depending on the predicted canopy height. We empirically find that linearly increasing the absolute threshold with height preserves the full range of predicted heights (Fig.~\ref{fig:regression_rh98}b), while causing a reduction of the overall error (Fig.~\ref{fig:uncertainty_vs_error}a). We therefore propose to filter according to $\hat{\sigma} < \tau (\hat{\mu} + \epsilon)$; i.e., a fixed minimal threshold $\epsilon$ is chosen for canopy height 0 empirically set to $\epsilon=10$~m. With increasing canopy height the cut-off value converges towards a relative threshold $\tau\hat{\mu}$ (appendix Fig.~\ref{fig:thresholds_filter_std}).
Ultimately, this filtering strategy reduces the underestimation bias (appendix Fig.~\ref{fig:boxplot_rh98_filtering}) and successfully discards the underestimated samples of the low power laser (coverage laser) in canopies\textgreater50~m (appendix Fig.~\ref{fig:boxplot_coverage_high_canopy}).

\subsection{Comparison to the first release of GEDI L2A height products}
The first release of GEDI L2A mission data contained estimates of height corresponding to six algorithm setting groups, each with a different combination of settings for signal and noise smoothing widths, and noise thresholds ~\citep{dubayah2020gedi_L2A}. These were designed to account for different observation conditions and waveform properties ~\citep{hofton_GEDI_ATBD}. 
\add{Prediction of optimal calibration settings of these algorithms is not available for the first GEDI L2A release, however improved calibration settings are planned for future releases. This reflects the iterative procedure towards waveform calibration: as more on-orbit data are processed, with increasingly better geolocation, calibration of waveform processing algorithms will be refined based not only on ALS cross-overs but also GEDI orbit cross-overs, eventually leading to final data products following mission completion with a single, best estimate of height (or some other derived parameter).
Nonetheless, the efficacy of our proposed CNN approach can be assessed through comparison with this first GEDI L2A release (Tab.~\ref{tab:comparison_gedi_l2a}).
}
On average, algorithm setting group 2 (out of the six possible) performs best over all continental regions with 4.4~m RMSE and $-$1.0~m ME, however it is important to note that the performance of individual algorithm setting groups changes regionally. 
If we always select the optimal of the six algorithms which is closest to the reference data, we find an RMSE of 3.4~m and an ME of $-$0.9~m.
Compared to this optimal selection, the CNN has slightly higher RMSE (3.6~m), but lower ME ($-$0.3~m) at 100~\% recall. Specifically, the CNN suffers less from underestimation of canopies \textless40~m but more so for canopies \textgreater40~m. 
Furthermore, our \rem{Bayesian approach}\addtwo{probabilistic approach} allows one to reduce the overall error by \addtwo{using the CNN output with the highest confidence, thus,} sacrificing some recall. When using only the 70~\% least uncertain samples, the RMSE drops to 2.7~m and the overall bias to $-$0.1~m ME, a significant improvement over the initial GEDI L2A height estimates, even when the optimal setting is \addtwo{hand}picked.

\begin{table*}[]
    \centering
    \begin{tabular}{lrr}
    \toprule
    {} &  RMSE [m] &  ME [m] \\
    \midrule
    GEDI L2A R001: Setting group 2 & 4.4 & -1.0  \\
    GEDI L2A R001: Optimal setting group & 3.4 & -0.9  \\
    CNN (100~\% recall) &       3.6 &       -0.3 \\
    CNN (70~\% recall) &       2.7 &       -0.1 \\
    \bottomrule
    \end{tabular}
    \caption{Comparison to the canopy top heights (RH98) from the first release (R001) of GEDI L2A products.}
    \label{tab:comparison_gedi_l2a}
\end{table*}

\subsection{Global canopy height predictions}

\begin{figure*}[t]
    \centering
    \subfloat[]{{\includegraphics[width=.98\textwidth]{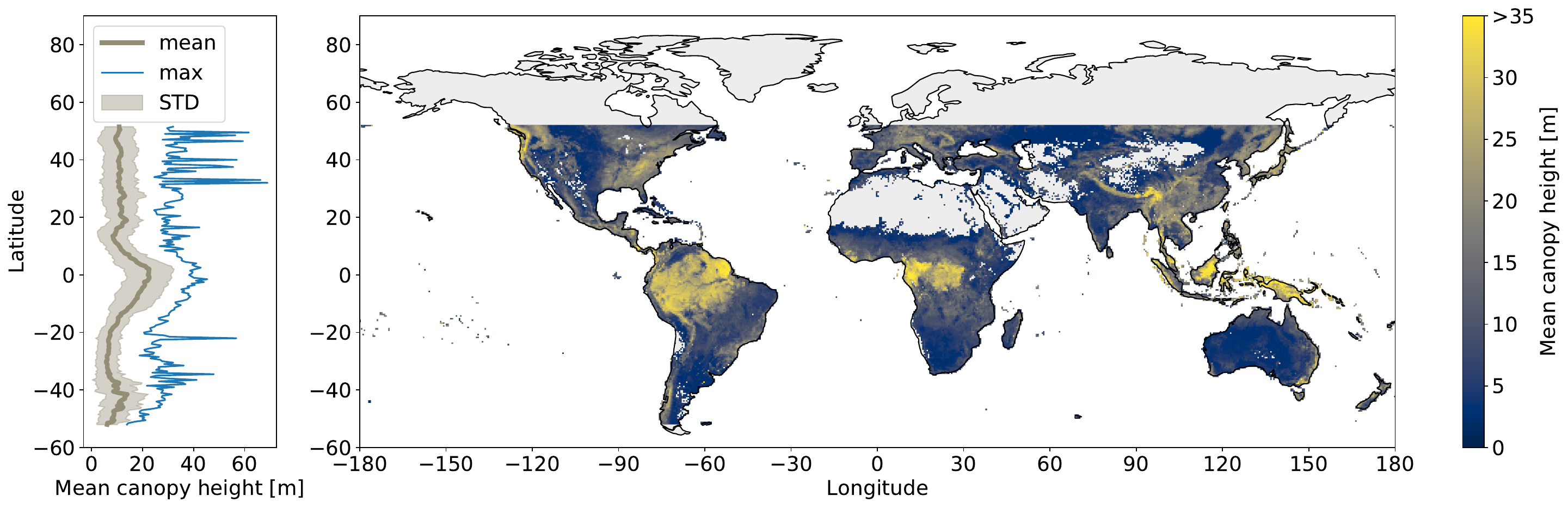}  }}%
    
    \subfloat[]{{\includegraphics[width=.98\textwidth]{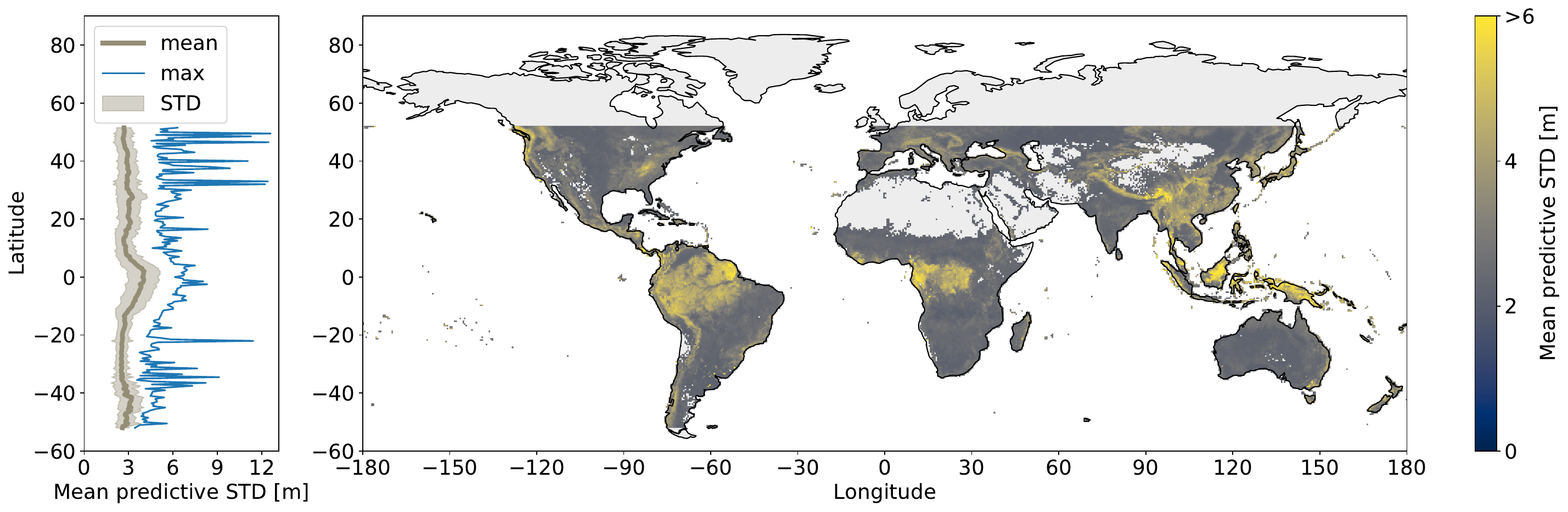} }}%
    
    \caption{Global canopy height predictions. (a) Mean canopy top height at 0.5 degree resolution ($\approx$55.5~km on the equator), created based on 346$\times$10\textsuperscript{6} waveform predictions from the first four months of the GEDI mission (April-July 2019). 
    \\(b) Mean predictive uncertainty (standard deviation, denoted as STD above) at the same raster resolution. On the left, the latitudinal distribution of mean and max heights, integrated around the globe.}
    \label{fig:global_prediction}
\end{figure*}

We illustrate a global canopy height map for the extent covered by GEDI (between 51.6 degrees latitude north and south) in Fig.~\ref{fig:global_prediction}. The map is computed from 1.8$\times$10\textsuperscript{9} L1B quality waveforms measured along 1,535 orbits during the first four months of the GEDI mission, from April to June 2019. We make use of the quality flag included in the corresponding L2A product, which is expected to yield waveforms with a good sensitivity of the ground returns. For the map, the individual heights per footprint have been gridded to 0.5~deg resolution, because of the sparse nature of GEDI observations within the four-month period. Final grid resolutions of 1~km or finer will be possible once data have been accumulated over two years.  
We apply three filtering steps before rasterizing canopy heights. First, we filter waveforms from non-vegetated areas, based on the "non-vegetated" layer from MODIS product MOD44B Version 6 Vegetation Continuous Fields~\citep{dimiceli2015mod44b}. We keep all samples with \textgreater70\% probability of being vegetation. This step is necessary to exclude urban areas, which would cause an increase in the RH98 and could deteriorate the large-scale canopy height patterns. 
Second, we filter based on the predictive uncertainty, with the setting that removed the 30\% most uncertain waveforms during random cross-validation (see above).
Third, we discard predictions with negative canopy height.
Alternatively, one could explicitly limit the mean output of all CNNs in the ensemble to zero, as it is done for the variance estimate. However, we prefer not to do this in favour of better epistemic uncertainty estimation, letting individual CNNs in the ensemble express their disagreement also for predictions close to 0~m.
Finally, we obtain canopy top height predictions for 346$\times$10\textsuperscript{6} footprints of 25~m diameter, ranging up to 97.2~m max canopy height. 
These sparse predictions are averaged over $\approx$55~km raster cells \add{(at the equator)} to obtain a global \add{dense} map (Fig.~\ref{fig:global_prediction}a). 

To study the geographical distribution of the predictive uncertainty, we average the individual predictive standard deviations into the same 0.5~degree ($\approx$55~km) raster (Fig.~\ref{fig:global_prediction}b). The corresponding maps for separated aleatoric and epistemic uncertainty are given in the appendix (Fig.~\ref{fig:global_uncertainty}).
As expected, the predictive uncertainty is highest in regions with high canopies. 
Finally, a closer look at the epistemic uncertainty (model uncertainty) at global scale reveals that the model, while able to predict canopy heights up to $\approx$95~m (footprint level, after filtering), is more uncertain
about predictions that exceed the 70~m height range of the training data (Fig.~\ref{fig:OOD_canopy_range} in the appendix).

\subsection{Potential to estimate ground elevation and other structure variables}
To demonstrate that \rem{Bayesian}\addtwo{deep ensembles of} convolutional neural networks can serve as a generic tool to process full waveform data, we present experiments in which, instead of canopy height, we regress two further parameters from the GEDI L1B waveforms. First, we estimate the relative height metric RH70, which provides a test of how well the method can predict structure within the canopy (that is below the top). Second, we estimate the ground elevation by regressing the elevation difference $z_{\Delta}=z_0 - z_g$ between the first waveform return $z_0$ and the true ground return $z_g$ (appendix Fig.~\ref{fig:regression_rh70} and Fig.~\ref{fig:regression_ground}).
Regression of RH70 leads to 2.2~m RMSE. This is significantly lower than for canopy top height (RH98), which is expected because RH70 values are in general lower, have a smaller range, and are less dependent on a well-defined, closed canopy top.
Estimation of the ground elevation yields an RMSE of 1.4~m. This error is lower compared to the RH98 regression primarily because remaining geolocation error in the reference dataset will have a lower impact on the ground heights, which are locally smooth, in contrast to the canopy top heights that can drastically change within a few meters. \addtwo{The impact of the geolocation error on the RH98 label noise is thus dependent on the heterogeneity of the vegetation~\citep{roy2021impact}.}

\section{Discussion}
Our proposed \rem{Bayesian CNN}\addtwo{probabilistic deep learning approach} provides a novel computational tool to estimate canopy top height from complex waveforms. It compares favorably with results from the first release of the GEDI L2A product (noting that this product is likely to improve considerably with subsequent releases). 
A major advantage of the CNN approach is that it avoids manual algorithm calibration to find the best settings, which can be a difficult and time-consuming process and often needs to be tuned to individual beams, plant functional types, canopy covers, and other observation conditions. The proposed end-to-end learning is data-driven, but generalizes across all possible scenarios, with the neural network encoding an adaptive waveform processor that is effective across these variable conditions. 
Traditional signal processing approaches to waveform interpretation exploit the near-direct measurement of height possible from LIDAR and are based on (and constrained by) physical principles. Whether or not the CNN ultimately out-performs traditional processing for near-direct measurements such as canopy height and cover remains to be seen, but our initial comparison with early GEDI products suggests this possibility.
Our results indicate that our data-driven approach may be a powerful alternative and potentially an important addition to the toolkit for space-borne LIDAR calibration. 

A key feature of our \rem{Bayesian CNN}\addtwo{approach} is that it provides direct estimation of predictive uncertainty, which, according to our evaluation, is well calibrated. This is a critical characteristic of our approach. 
It can be used to filter canopy height predictions to meet specific accuracy requirements for a given application. By design, GEDI provides only a sample of the Earth's surface, with calibrated uncertainties, the trade-off between reduced sample size (recall) and precision can be explicitly balanced. Thus, canopy height may be filtered with an adaptive threshold, to reduce predictive uncertainty to the degree needed to preserve the range of estimated heights and their geographical coverage, as illustrated by our experiments.
\add{The estimated predictive uncertainty does not include uncertainties introduced by systematic geolocation errors when applied to uncorrected waveforms, but will implicitly capture label noise introduced by the remaining geolocation errors in the collocated training dataset.} 

Our model is able to predict canopy top heights up to 97.2~m, which means that it is able cover the relevant height range for the vast majority of forests on Earth~\citep{tao2016global}, and to make predictions that exceed the range of canopy heights contained in the training dataset (in our case up to 70.1~m). As expected, the epistemic uncertainty is higher for canopies above 70~m across the globe and we observed a systematic underestimation of tall canopy heights.
We note that the tallest trees generally have the highest biomass, and therefore a systematic underestimation of the associated biomass may have implications for carbon cycle applications.
Since the model is trained by minimizing the overall error, our assumption is that this bias is caused by the dataset sampling bias (see the imbalanced distribution of RH98 in Fig.~\ref{fig:gt_data}b). Hence, collecting more training data with tall canopy heights will decrease both uncertainty and bias. 
\add{However, the over- and underestimation in the lower and upper range, respectively, can also be caused by the tendency to predict values towards the mean of the training data if the input waveform is noisy and thus less informative.}
In general, the epistemic uncertainty may guide the collection of new training data that is most informative to the model.
Compensating such dataset biases in the training procedure is an active, yet unsolved, research topic~\citep{cui2019class}. It remains to be investigated how methods developed particularly for classification can be transferred to regression.
Nevertheless, more on-orbit data over the available ALS reference regions will allow to create more training samples, especially in the tropics where the sample density is currently sparser than in high latitudes due to the ISS orbit and the frequent cloud coverage.

GEDI utilizes a coverage laser that is optically split into two beams to improve coverage (each power laser produces two tracks of footprints, while the coverage laser produces four tracks)~\citep{DUBAYAH2020100002}. Due to the reduced power, the coverage beams exhibit lower signal-to-noise ratio, which makes it harder to detect weak returns from both the ground and the canopy. This difference in power did not affect overall CNN performance, which is an encouraging result that might pave the way towards a less restrictive use of GEDI's coverage beams. For example, in its current biomass estimation protocol, GEDI does not use coverage beams during daylight conditions for canopy cover exceeding 70~\% because of weak ground returns~\citep{DUBAYAH2020100002}. Additionally, maximizing the accuracy and usability of all GEDI data has important implications for derived map products that aggregate the footprint predictions, e.g., to 1~km grid cells.
Reliable footprint estimates will also advance downstream mapping tasks based on multi-modal data fusion, e.g., combining sparse GEDI canopy height predictions with \add{wall-to-wall raster} satellite data from other missions like Sentinel-2~\citep{lang2019country}, TanDEM-X~\citep{lee2018gedi,qi2019improved}, or Landsat~\citep{healey2020highly,potapov2020mapping}.

Another useful property of the generic, data-driven CNN approach to waveform processing is that it can be directly applied to other variables one wishes to derive from the raw waveform, such as ground elevation, plant area index, but also biophysical variables that are often correlated with the LIDAR returns, such as biomass or species richness. For instance, most models that predict biomass from LIDAR use a variety of derived waveform metrics and relate them to ground-based biomass estimates through some form of regression. It may be possible to bypass the explicit computation of waveform predictor metrics, such as canopy height, and their associated predictive uncertainty, and instead regress biomass end-to-end from the raw waveform, potentially uncovering more intricate relationships between waveform morphometry and biomass in a data-driven manner. 

\section{Conclusion}
In conclusion, the proposed \rem{Bayesian CNN model}\addtwo{probabilistic deep learning approach} has demonstrated the ability to navigate the complex task of deriving global canopy heights from on-orbit L1B GEDI waveforms with favorable accuracies that may exceed traditional waveform processing. Formulating canopy height regression from GEDI waveforms as an end-to-end supervised learning task avoids the explicit modelling of unknown atmospheric, environmental, and instrument effects and instead learns to directly isolate the desired signal from the noisy data. Moreover, our experiments have confirmed that the CNN model learns to extract robust features that generalize to unseen geographical regions \add{that are not included in the training dataset}. Modelling the predictive uncertainty with a \rem{Bayesian}\addtwo{probabilistic} learning strategy supplements the canopy height regression with uncertainty estimates that, empirically, are correctly calibrated and reliable. These estimates of predictive uncertainty, among other uses, make it possible to filter out highly uncertain predictions, which reduces the empirical error, while preserving the full range of canopy heights and their global distribution. 


\section{Data availability}
The generated canopy height map data are available as georeferenced raster files from \url{https://doi.org/10.5281/zenodo.5112903} \citep{lang_nico_2021_5112904}.

Information on the ALS derived reference data that support the findings of this study are available on request from the GEDI mission science team. Requests must be directed to John Armston (armston@umd.edu) or Ralph Dubayah (dubayah@umd.edu). Most of the source ALS data are open access. ALS data from the Terrestrial Ecosystem Research Network in Australia are available from \url{http://www.auscover.org.au/purl/airborne-lidar-qa-all-sites}. ALS data from the NASA Carbon Monitoring System (CMS) on Borneo Island, Kalimantan, Indonesia, are available from \url{https://doi.org/10.3334/ORNLDAAC/1518}. ALS data from the National Ecological Observatory Network (NEON) in the United States are available from \url{https://data.neonscience.org/data-products/explore}. Information on non-public ALS datasets are in the acknowledgments and subject to data use agreements.

\section{Code availability}
The code to train and deploy the convolutional neural network is available at: \url{https://github.com/langnico/GEDI-BDL}.

\section{Acknowledgements}
We thank the GEDI mission science team, in particular Carlos Edibaldo Silva, Laura Duncanson, Steven Hancock and David Minor for their contributions to the collation and processing of the ALS reference dataset, and Michelle Hofton and Bryan J. Blair for guidance on the GEDI Level 2A algorithm.
Eliakimu Zahabu from the Sokoine University of Agriculture (SUA), Erik N{\ae}sset and Terje Gobakken from the Norwegian University of Life Sciences (NMBU) gave permission to use ALS data in Liwale and Amani Nature Reserve (Tanzania). These data were acquired by financial contribution from the Royal Norwegian Embassy in Tanzania under the grant entitled "Enhancing the measuring, reporting and verification (MRV) of forest in Tanzania through the application of advanced remote sensing techniques". Hans Verbeeck and Elizabeth Kearsley (University of Ghent) gave permission to use ALS data acquired at the Yangambi in the Democratic Republic of the Congo. Ross Hill (Bournemouth University) gave permission to use ALS data acquired in the New Forest (United Kingdom). Sassan Saatchi (NASA JPL) granted access to the ALS data over Lope in Gabon, which will eventually be released through SilvaCarbon after an embargo period.
The project received funding from Barry Callebaut Sourcing AG, as part of a Research Project Agreement, and funding from NASA contract {\#}NNL 15AA03C to the University of Maryland for the development and execution of the GEDI mission (Dubayah, Principal Investigator).



\clearpage
\bibliography{bibliography}

\clearpage
\onecolumn
\appendix

\newpage
\section{Computation time}
Scaling to the large data volume of a global product is possible thanks to efficient parallel computing on modern graphics processing units (GPUs). The computation time for a single orbit file with 1.2$\times$10\textsuperscript{6} waveforms is \textless10 minutes for the complete model ensemble, on a single consumer-grade GPU (Nvidia GeForce GTX 1080Ti). Predicting canopy height for our archive of four months of quality GEDI L1B waveforms (12.3~TB of raw waveform data) amounts to approximately 250 hours of GPU computing, where every waveform is processed independently, so that the process can easily be parallelized on a GPU cluster to reduce effective computation time.

\section{CNN architecture illustration}

\begin{figure*}[th]
    \centering
    \includegraphics[width=\textwidth]{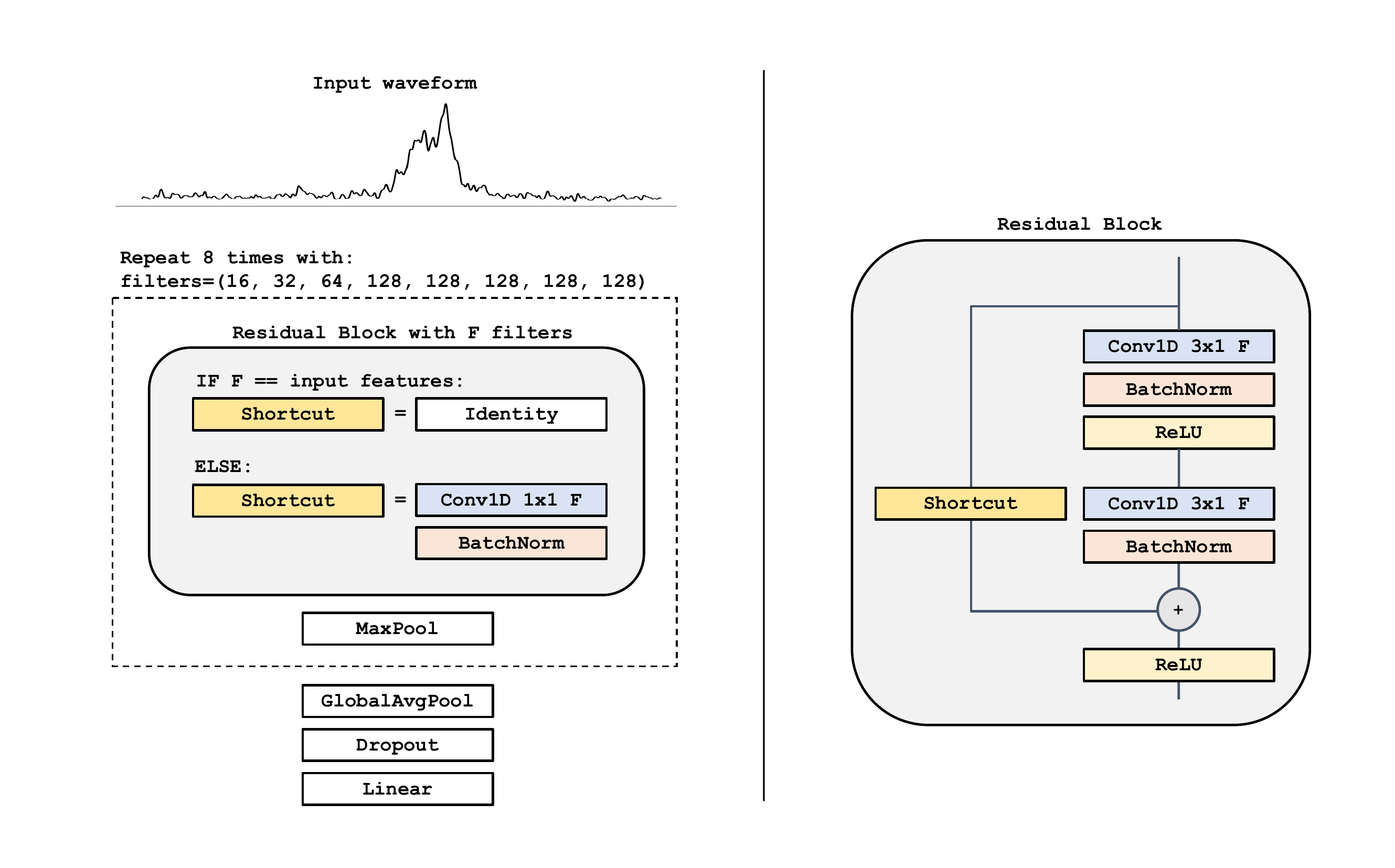}
    \caption{Graphical illustration of the convolutional neural network based on the ResNet architecture.}
    \label{fig:CNN_illustration}
\end{figure*}

\newpage
\section{Performance regarding laser power for high canopies}
\begin{figure*}[th]
    \centering
    \subfloat[]{{\includegraphics[width=0.45\textwidth]{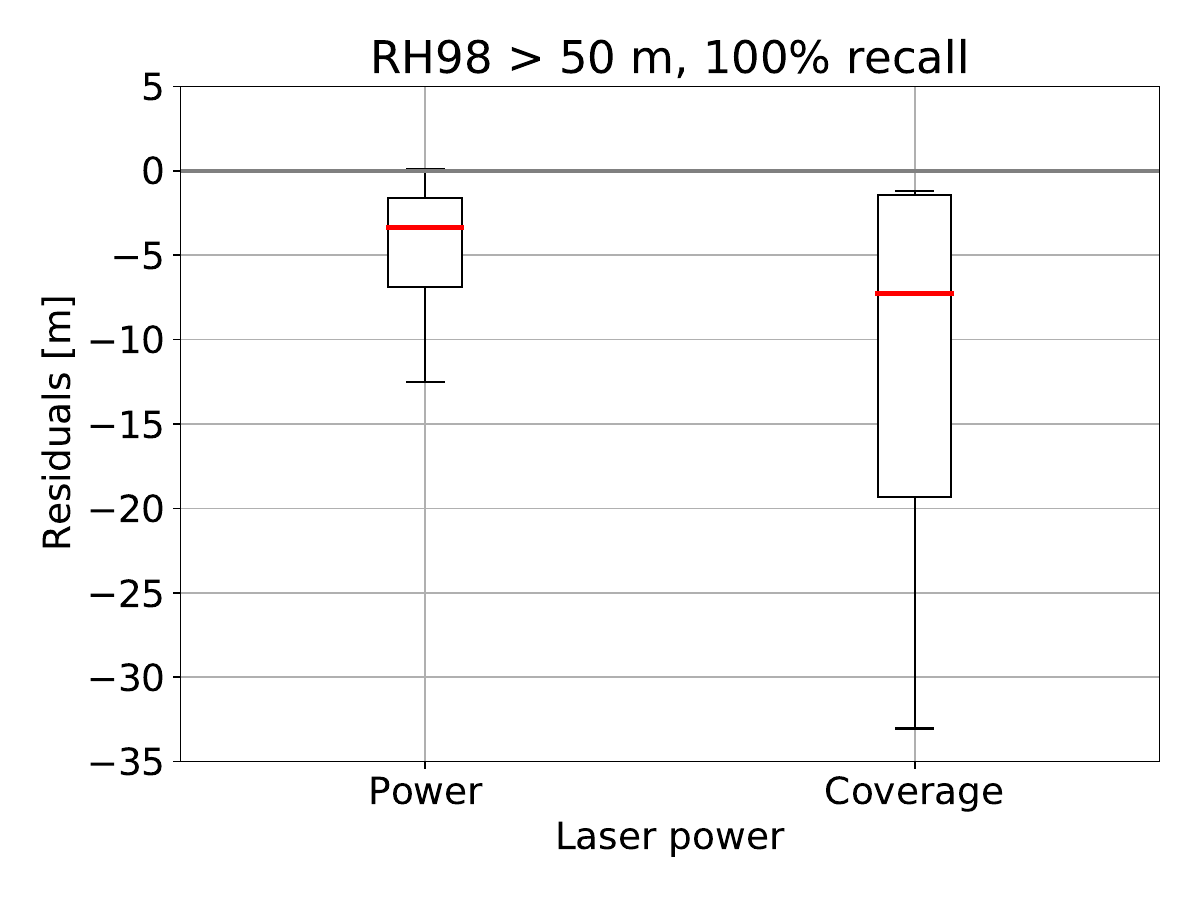} }}%
    \qquad
    \subfloat[]{{\includegraphics[width=0.45\textwidth]{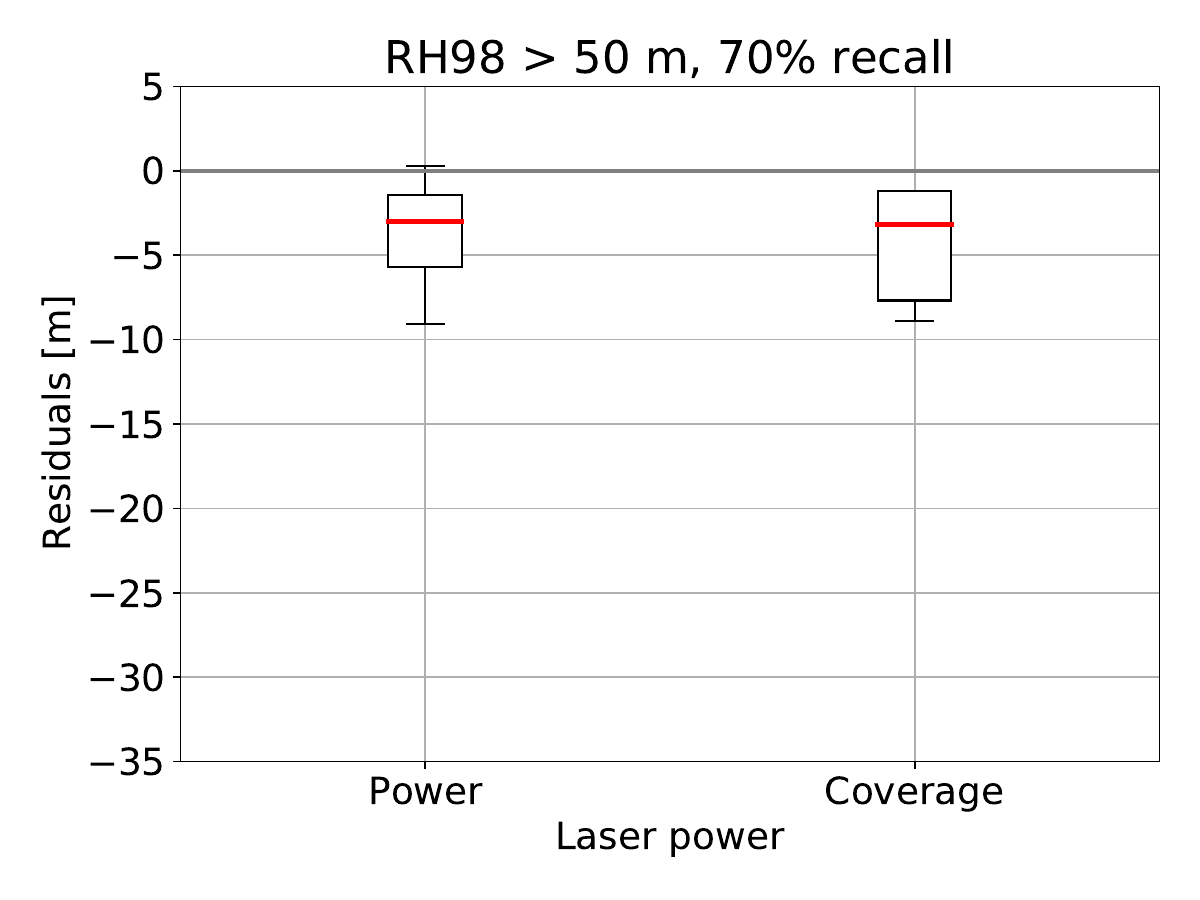} }}%
    \caption{Performance with respect to laser power for samples with canopy height \textgreater50~m (random cross-validation results). Evaluation over (a) all test samples (100\% recall) and (b) filtered test samples with only the 70\% most certain predictions.}
    \label{fig:boxplot_coverage_high_canopy}
\end{figure*}

\section{Performance regarding solar background noise and beam sensitivity}
\begin{figure*}[th]
    \centering

    \subfloat[]{{\includegraphics[width=0.45\textwidth]{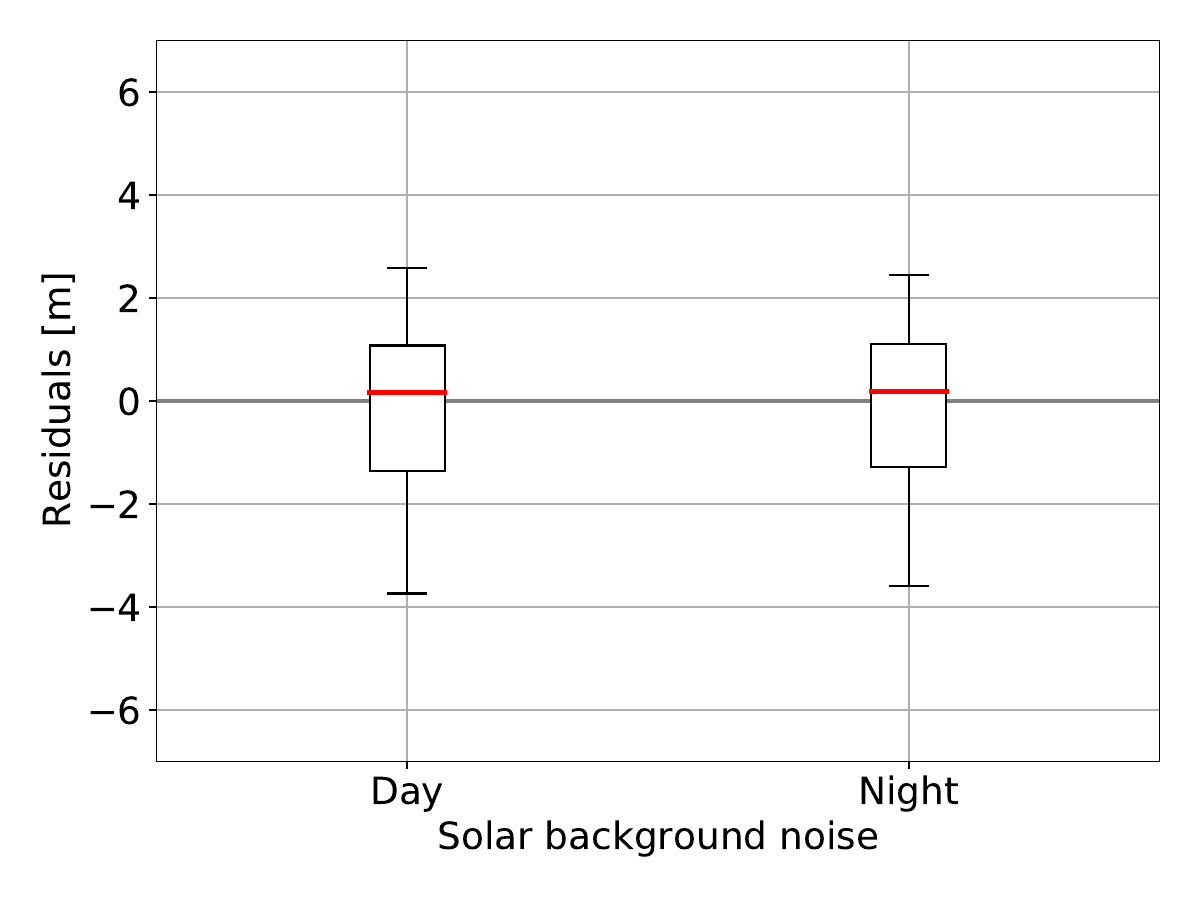} }}%
    \qquad
    \subfloat[]{{\includegraphics[width=0.45\textwidth]{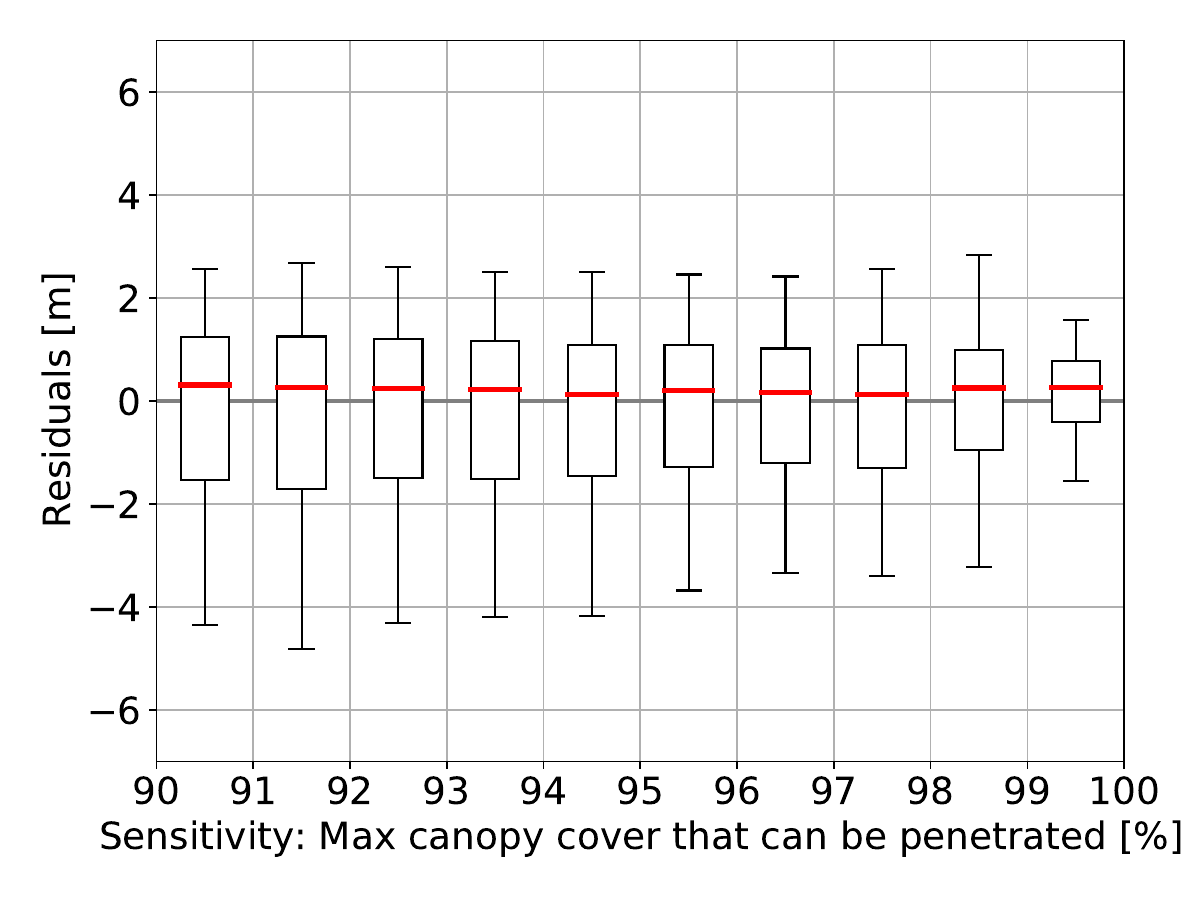} }}%
    
    \caption{(a) Performance regarding solar background noise (random cross-validation results). (b) Performance regarding beam sensitivity, i.e., the maximum canopy cover that can be penetrated.}
    \label{fig:boxplot_solar_and_sensitivity}
\end{figure*}

\newpage
\section{Geographic generalization}
\begin{figure*}[th]
    \centering
    \subfloat[Europe]{{\includegraphics[width=0.45\textwidth]{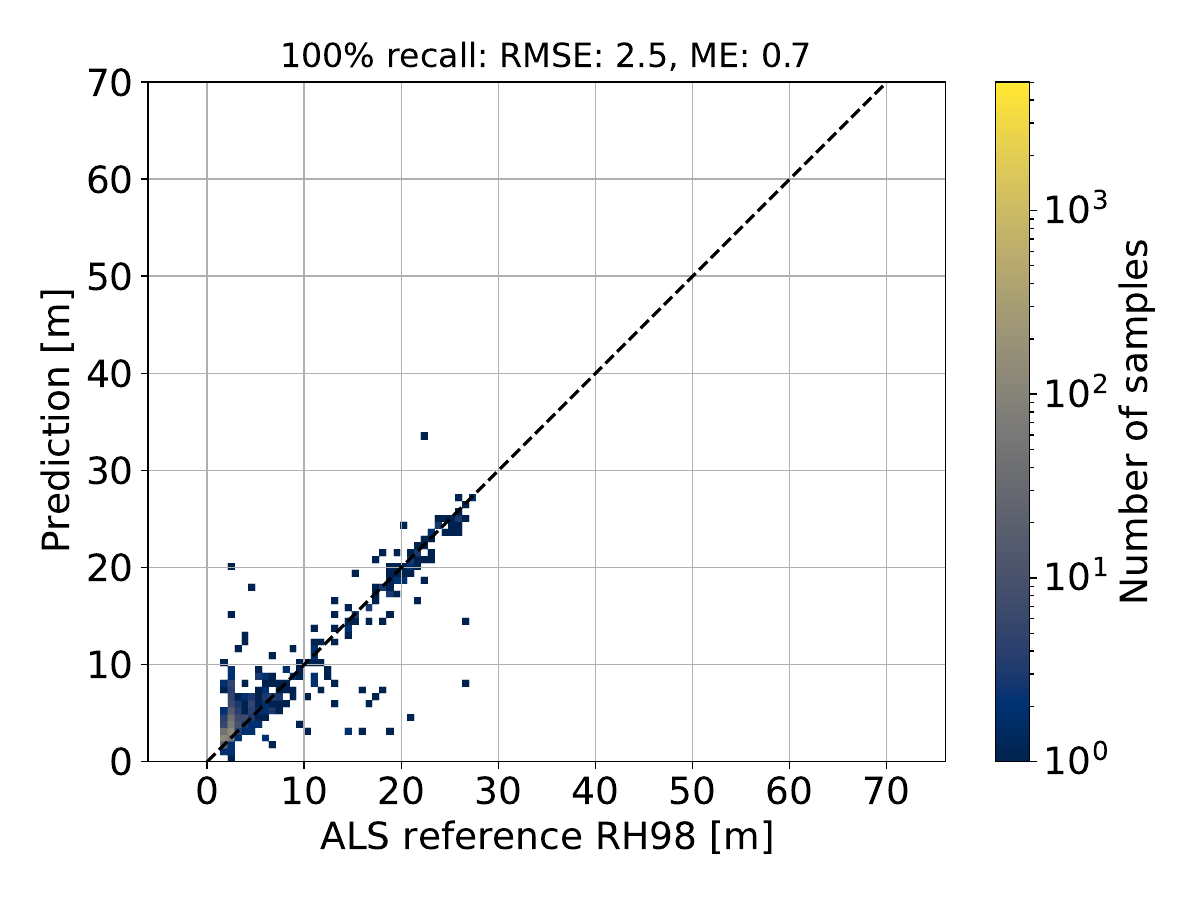} }}%
    \subfloat[Australia]{{\includegraphics[width=0.45\textwidth]{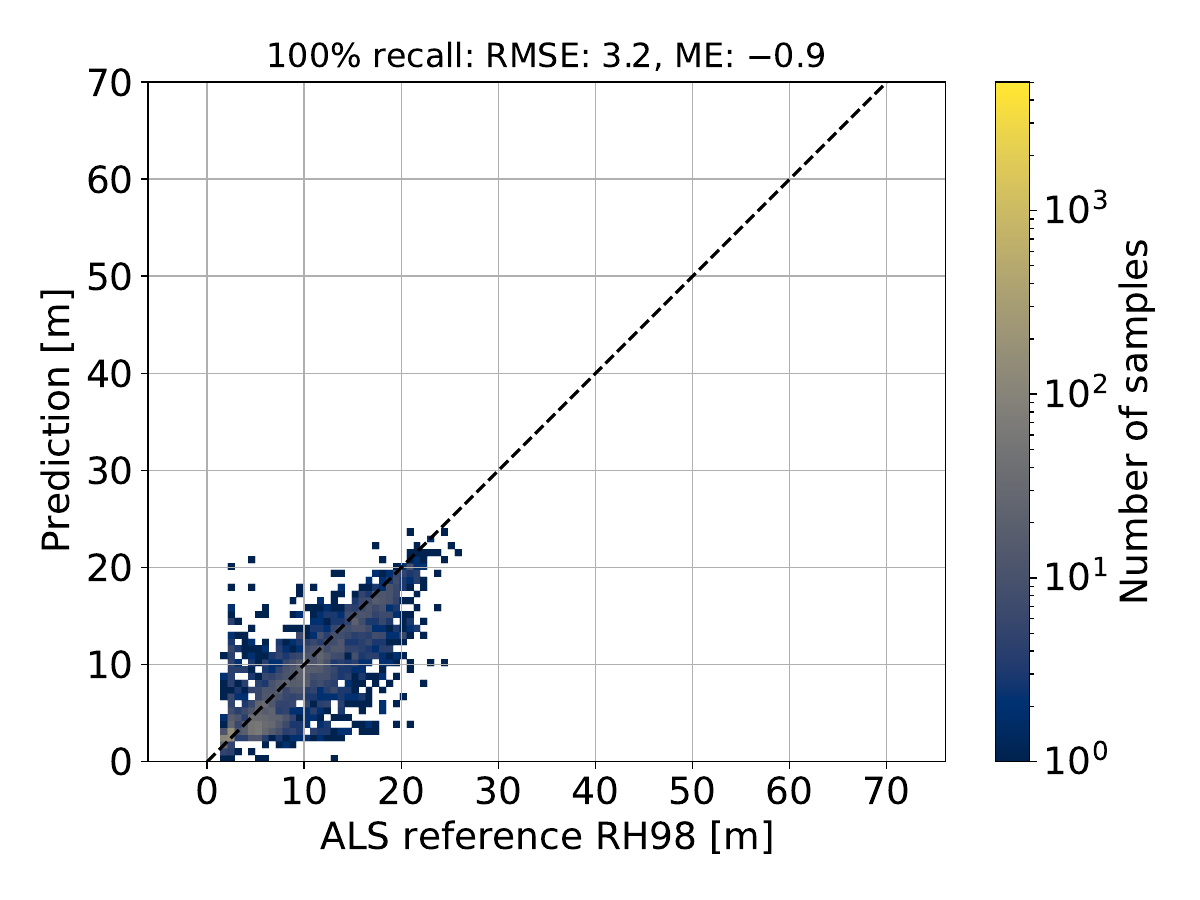} }}%

    \subfloat[Africa]{{\includegraphics[width=0.45\textwidth]{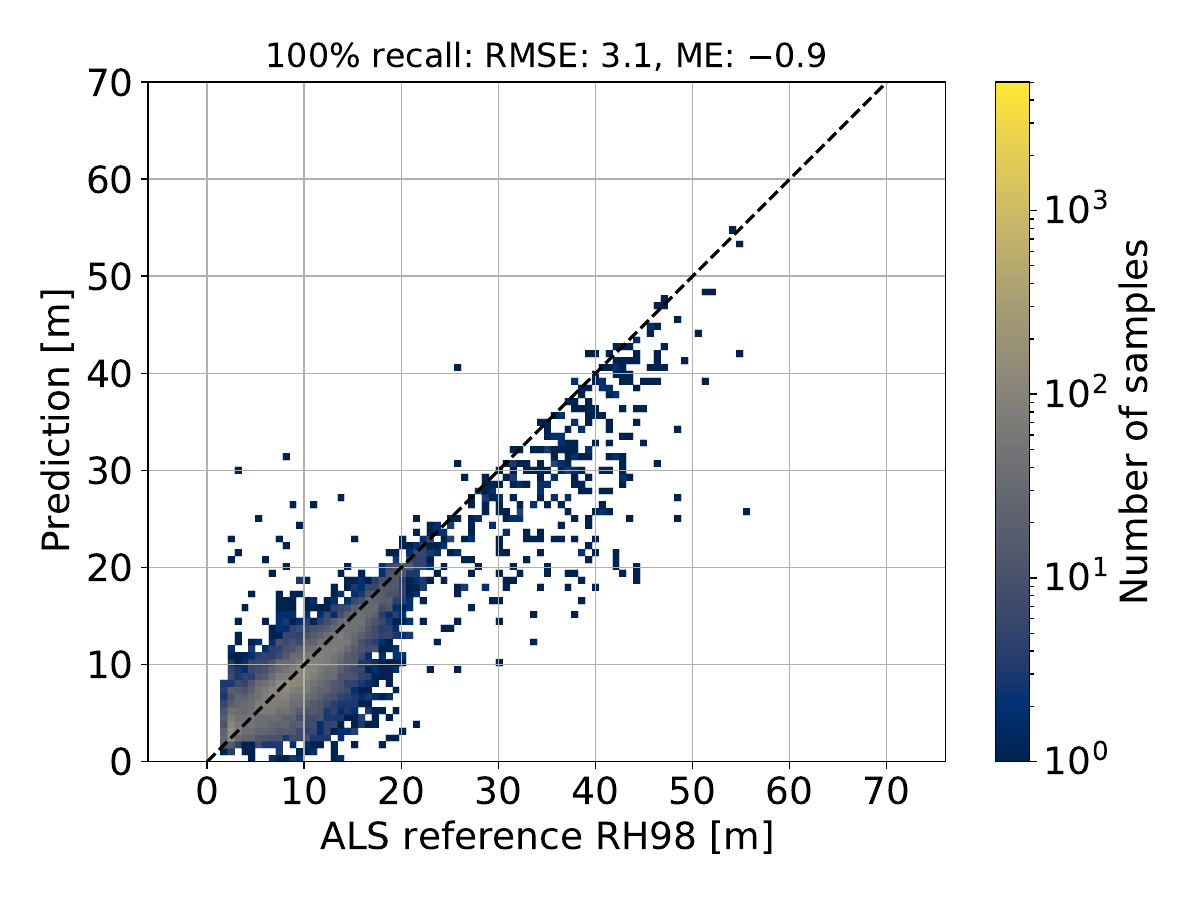} }}%
    \subfloat[Tropics]{{\includegraphics[width=0.45\textwidth]{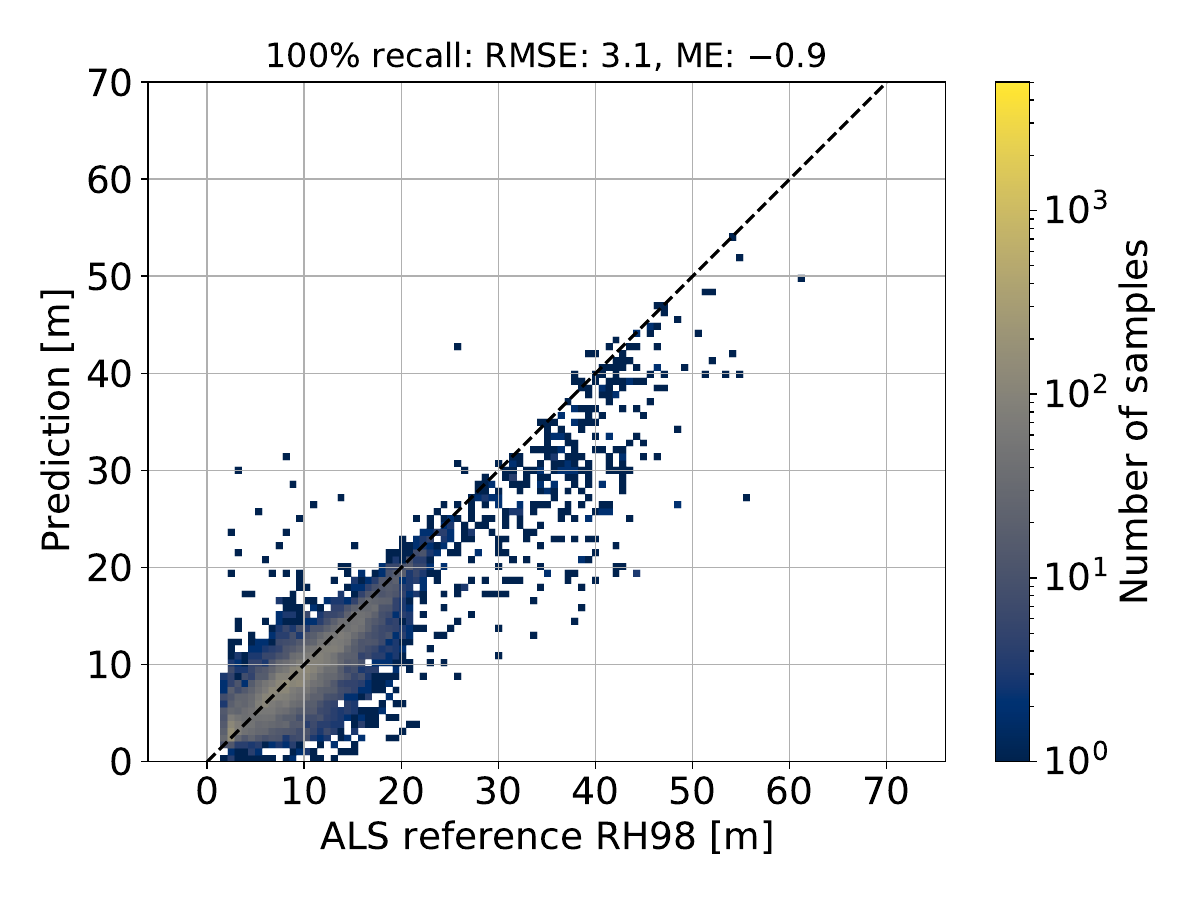} }}%
    \caption{Geographic generalization experiments regressing canopy top height (RH98).}
    \label{fig:geographic_generalization}
\end{figure*}

\newpage
\section{Filtering with absolute and relative predictive uncertainty thresholds}
\begin{figure*}[th]
    \centering
    \subfloat[]{{\includegraphics[width=0.45\textwidth]{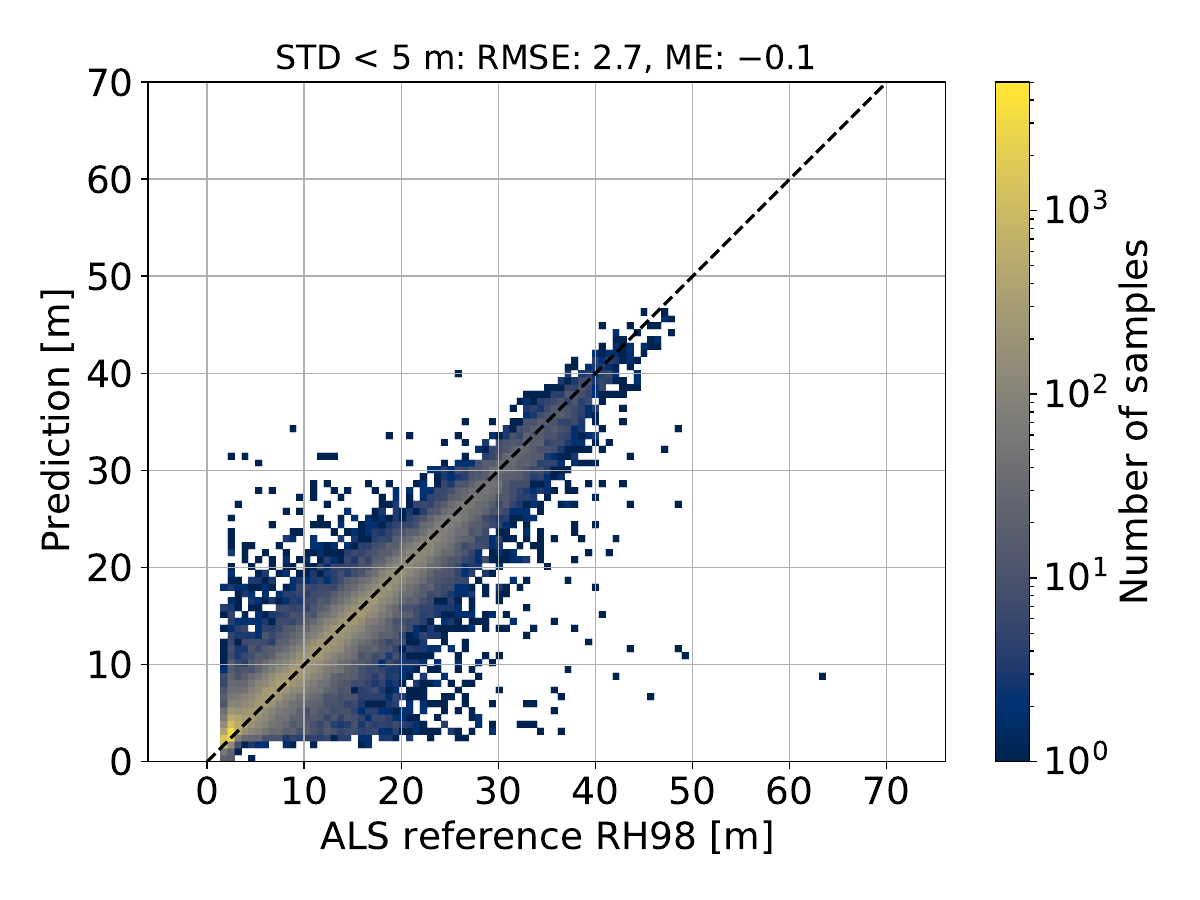} }}%
    \subfloat[]{{\includegraphics[width=0.45\textwidth]{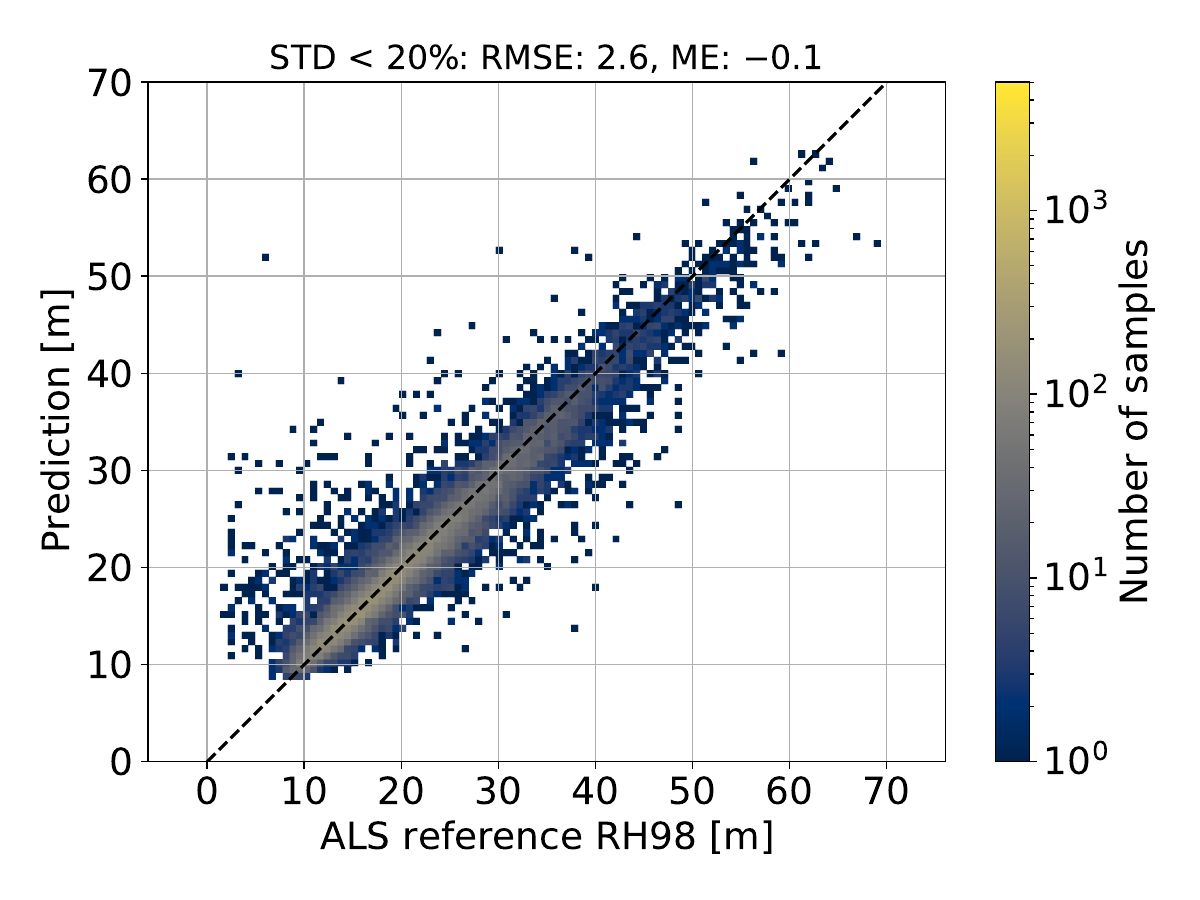} }}%

    \caption{Regression of canopy top height (RH98) from L1B GEDI full waveforms. ALS reference vs.\ CNN estimates, with different filtering settings: (a) Test samples with absolute predictive standard deviation \textless5~m. (b) Test samples with relative predictive standard deviation \textless20\%.}
    \label{fig:filtering_abs_rel}
\end{figure*}

\section{Filtering with predictive uncertainty using adaptive thresholds}

\begin{figure*}[th]
    \centering
    \subfloat[]{{\includegraphics[width=0.45\textwidth]{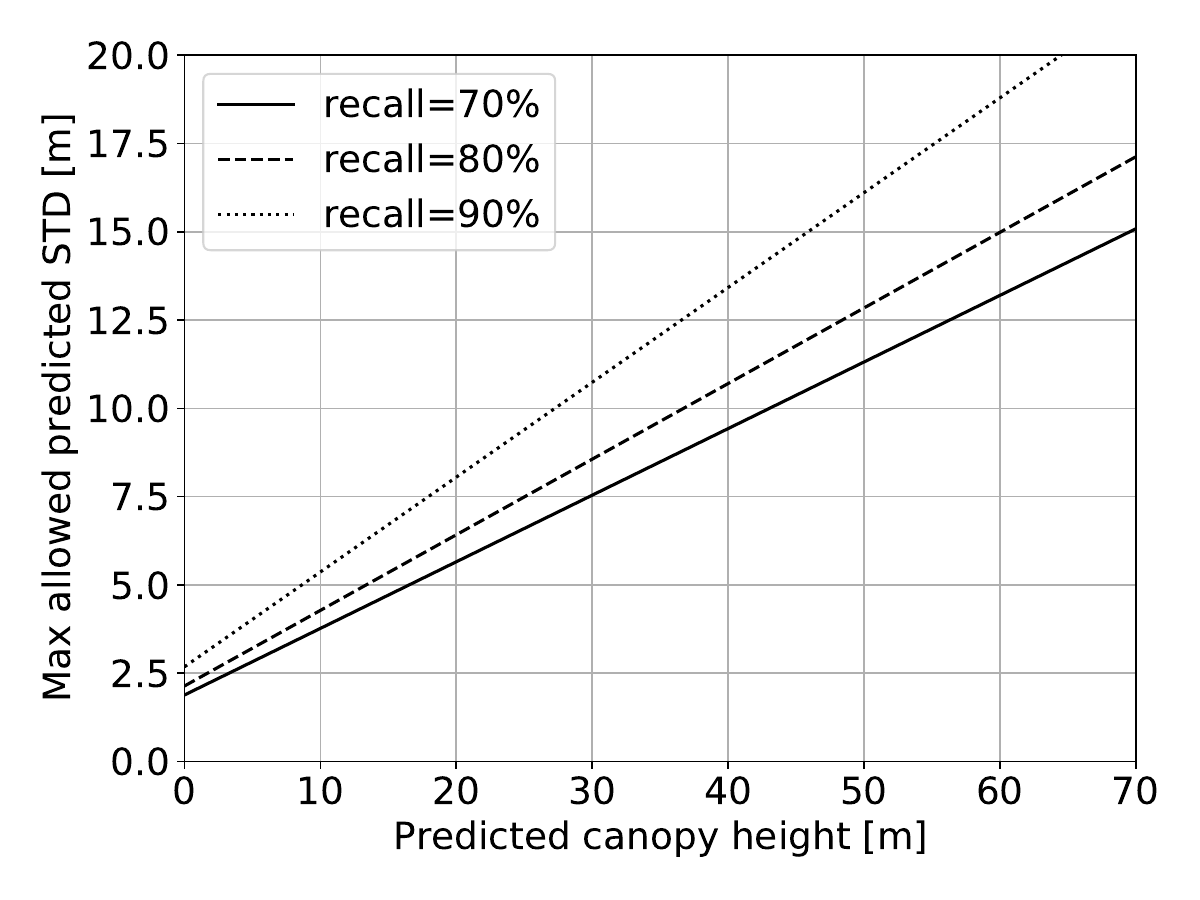} }}%
    \qquad
    \subfloat[]{{\includegraphics[width=0.45\textwidth]{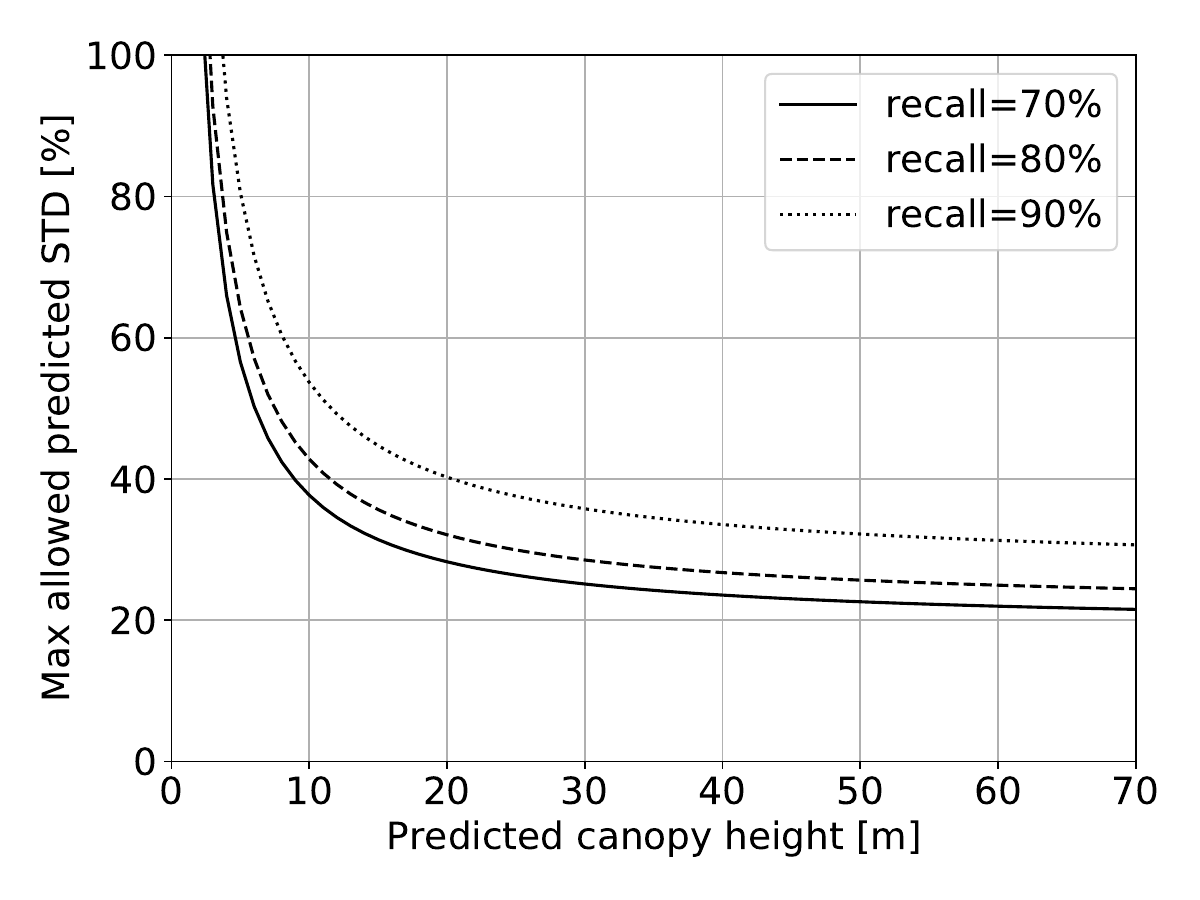} }}%
    \caption{
   Adaptive thresholds for filtering predictions. The diagrams show the numerical values of the threshold function $\tau(\hat{\mu}+\epsilon)$ for canopy heights 0-70~m (left absolute, right relative).
The adaptive thresholding approach preserves the full range of canopy heights.}%
    \label{fig:thresholds_filter_std}%
\end{figure*}

\newpage
\section{Effect of the filtering with predictive uncertainty}
\begin{figure*}[th]
    \centering
    \subfloat[]{{\includegraphics[width=0.45\textwidth]{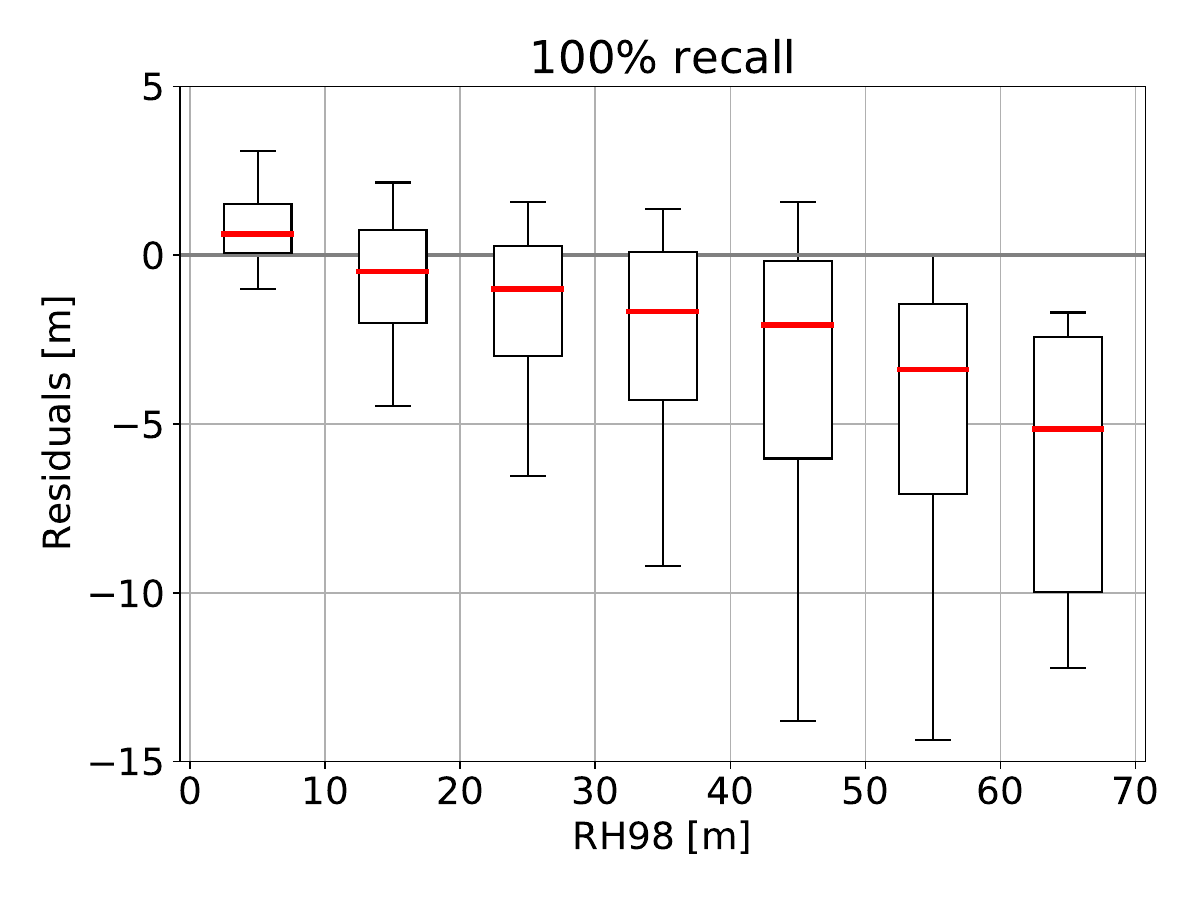} }}%
    \qquad
    \subfloat[]{{\includegraphics[width=0.45\textwidth]{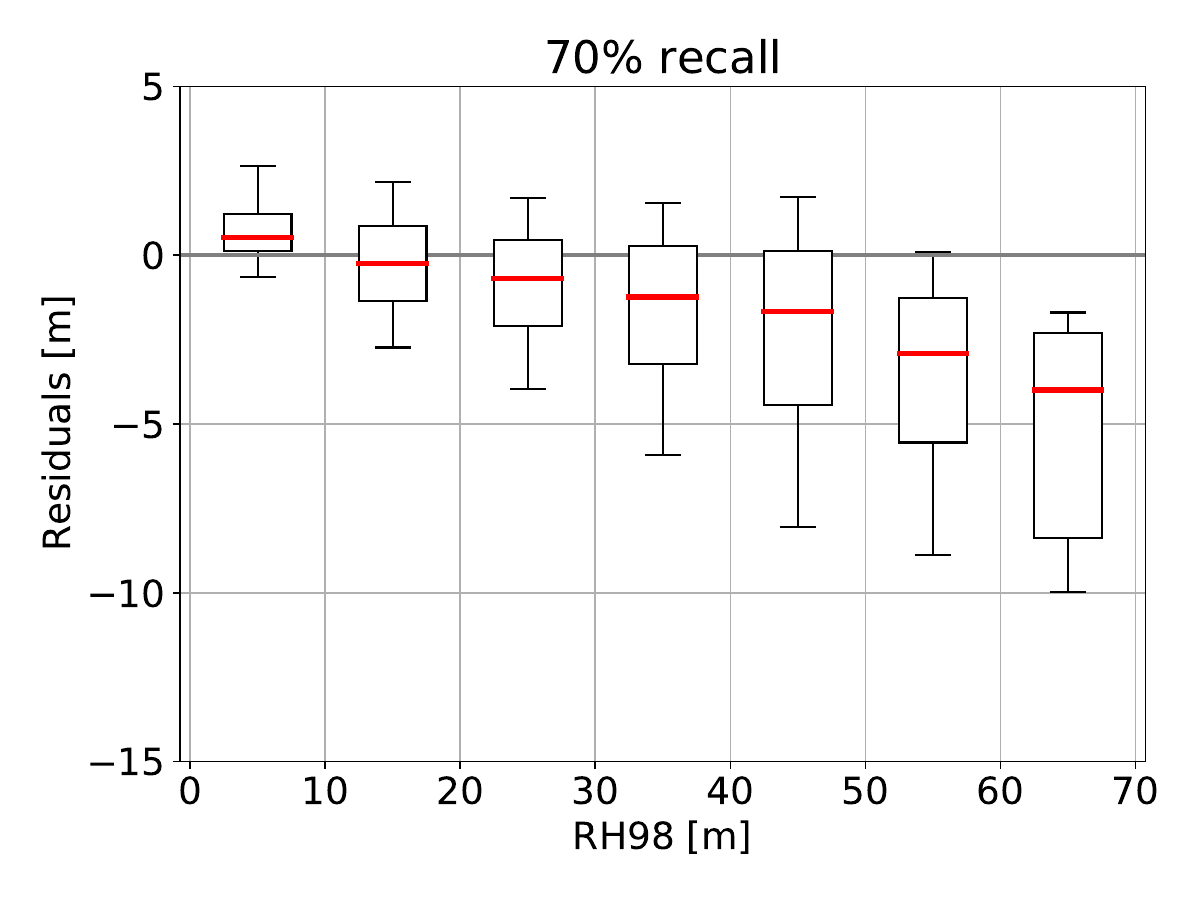} }}%
    \caption{Performance with respect to RH98 (random cross-validation results). Evaluation over (a) all test samples (100\% recall) and (b) filtered test samples with only the 70\% most certain predictions.}
    \label{fig:boxplot_rh98_filtering}
\end{figure*}

\section{Calibration of aleatoric and epistemic uncertainty only}
\begin{figure*}[th]
    \centering
    \subfloat[]{{\includegraphics[width=0.45\textwidth]{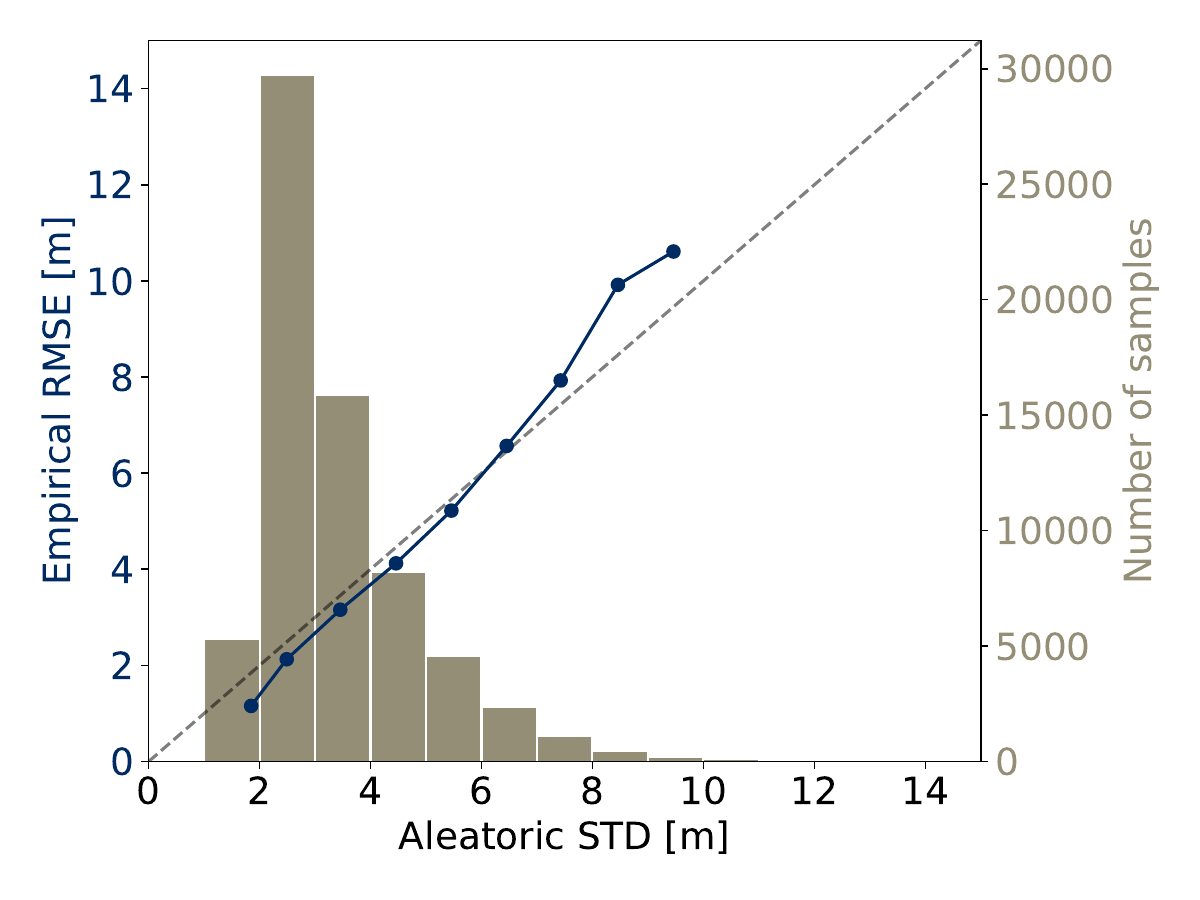} }}%
    \qquad
    \subfloat[]{{\includegraphics[width=0.45\textwidth]{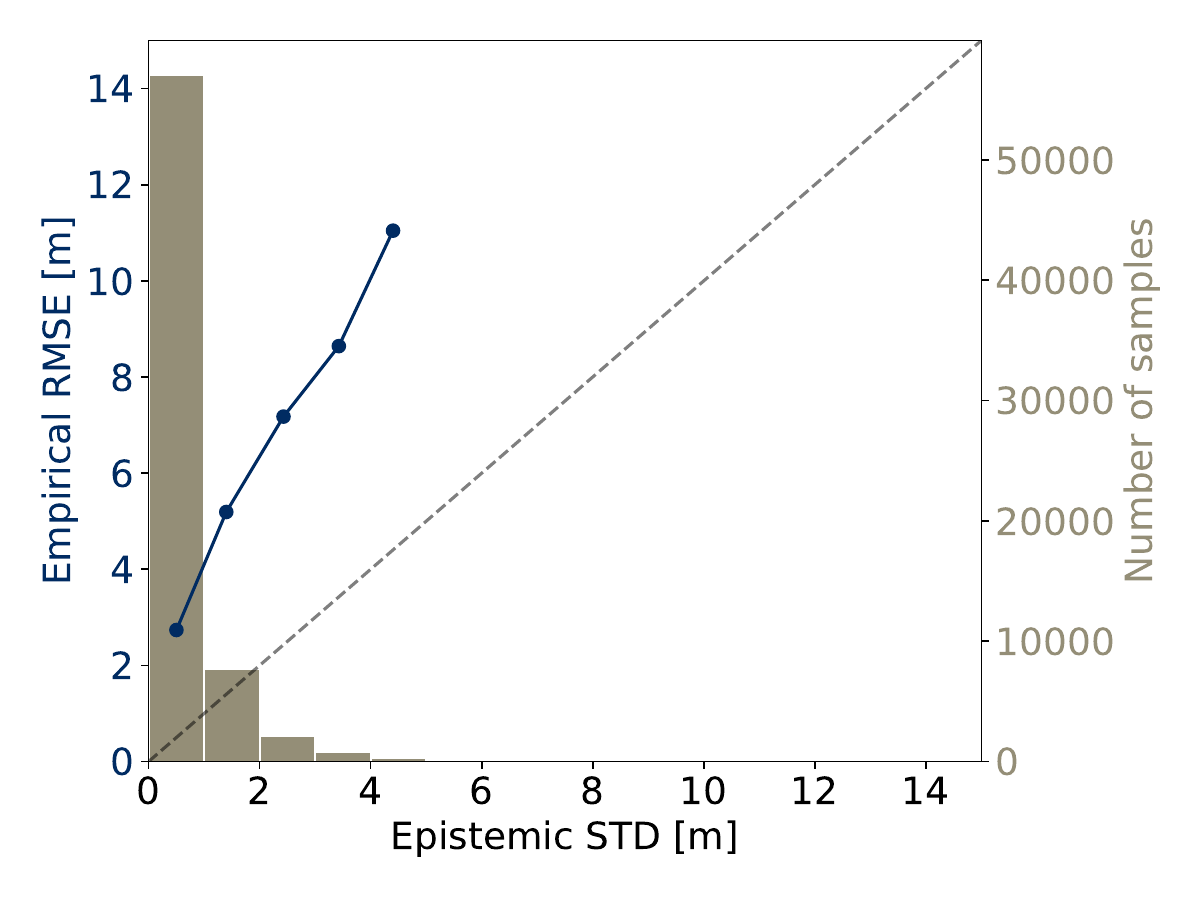} }}%
    \caption{Calibration plots for the random cross-validation of RH98 using (a) the aleatoric uncertainty and (b) the epistemic uncertainty. The estimated standard deviation (STD) is plotted vs.\ the empirical RMSE for subsets grouped by the standard deviation.}
    \label{fig:calibration_plots}
\end{figure*}

\newpage
\section{Global aleatoric and epistemic uncertainty maps}

\begin{figure*}[thb]
    \centering
    \subfloat[]{{\includegraphics[width=.98\textwidth]{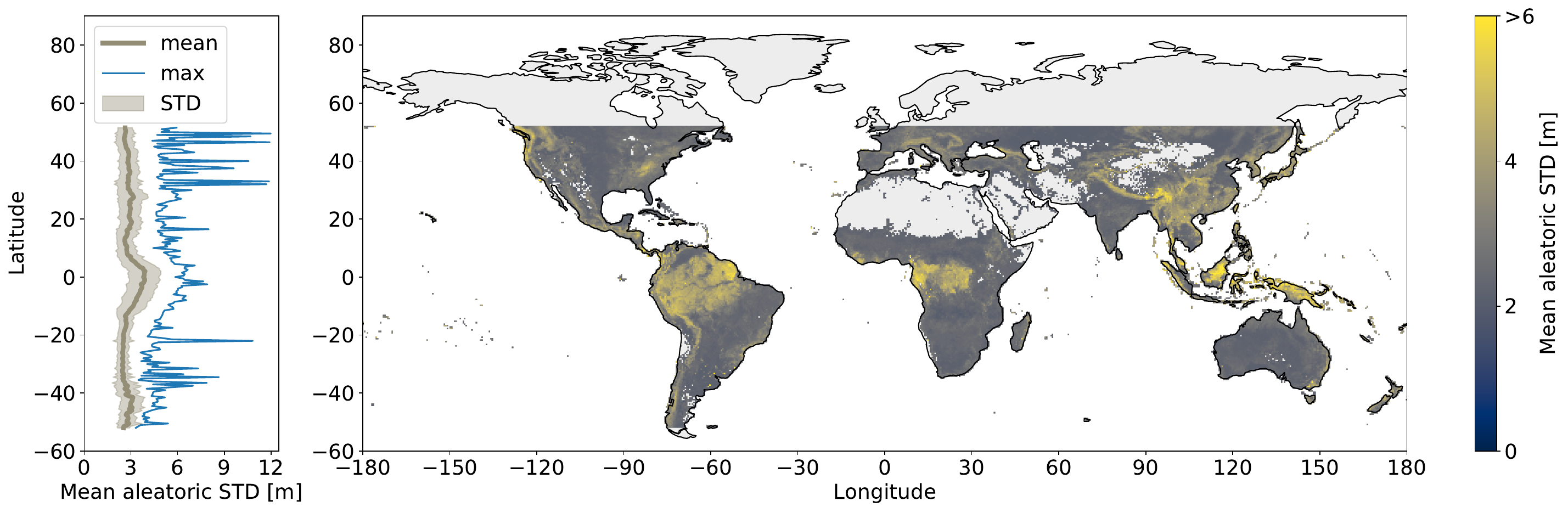}  }}%
    
    \subfloat[]{{\includegraphics[width=.98\textwidth]{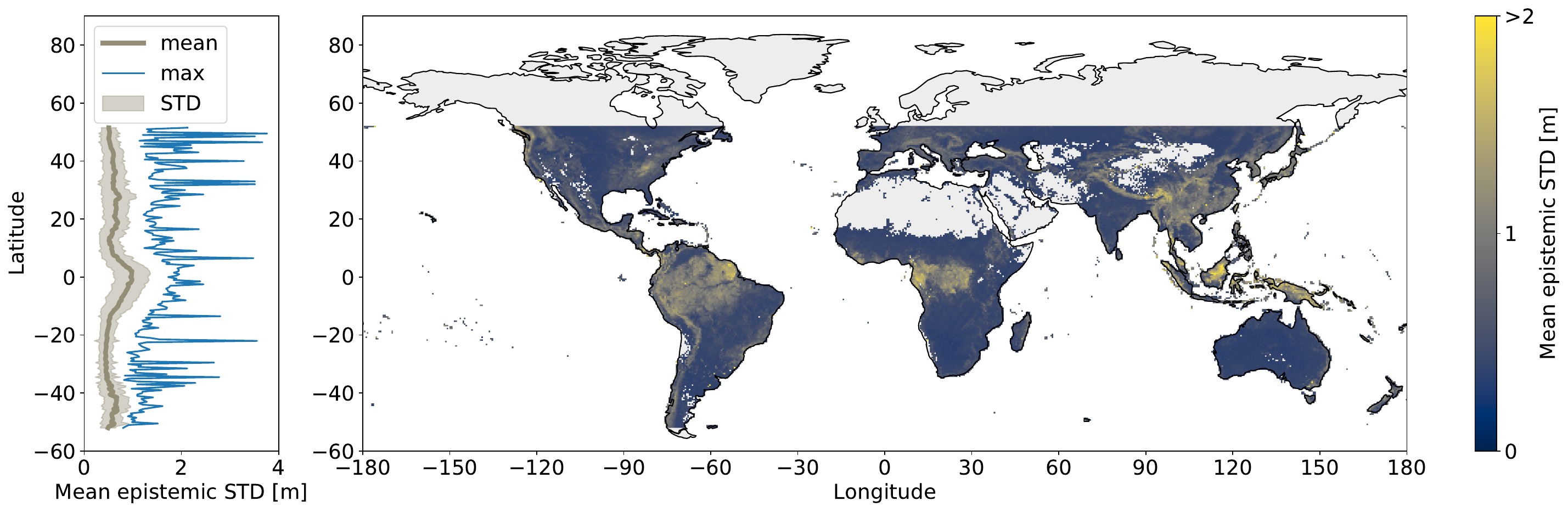} }}%
    
    \caption{Global uncertainty maps for the canopy height predictions. (a) Mean aleatoric uncertainty (standard deviation, denoted as STD above) at 0.5 degree resolution ($\approx$55.5~km on the equator), created based on 346$\times$10\textsuperscript{6} waveform predictions from the first four months of the GEDI mission (April-July 2019). 
    (b) Mean epistemic uncertainty at the same raster resolution. On the left, the latitudinal distribution of mean and max heights, integrated around the globe.}
    \label{fig:global_uncertainty}
\end{figure*}

\newpage
\section{Global model (epistemic) uncertainty regarding canopy top height}

\begin{figure*}[h]
    \centering
   \includegraphics[width=0.45\textwidth]{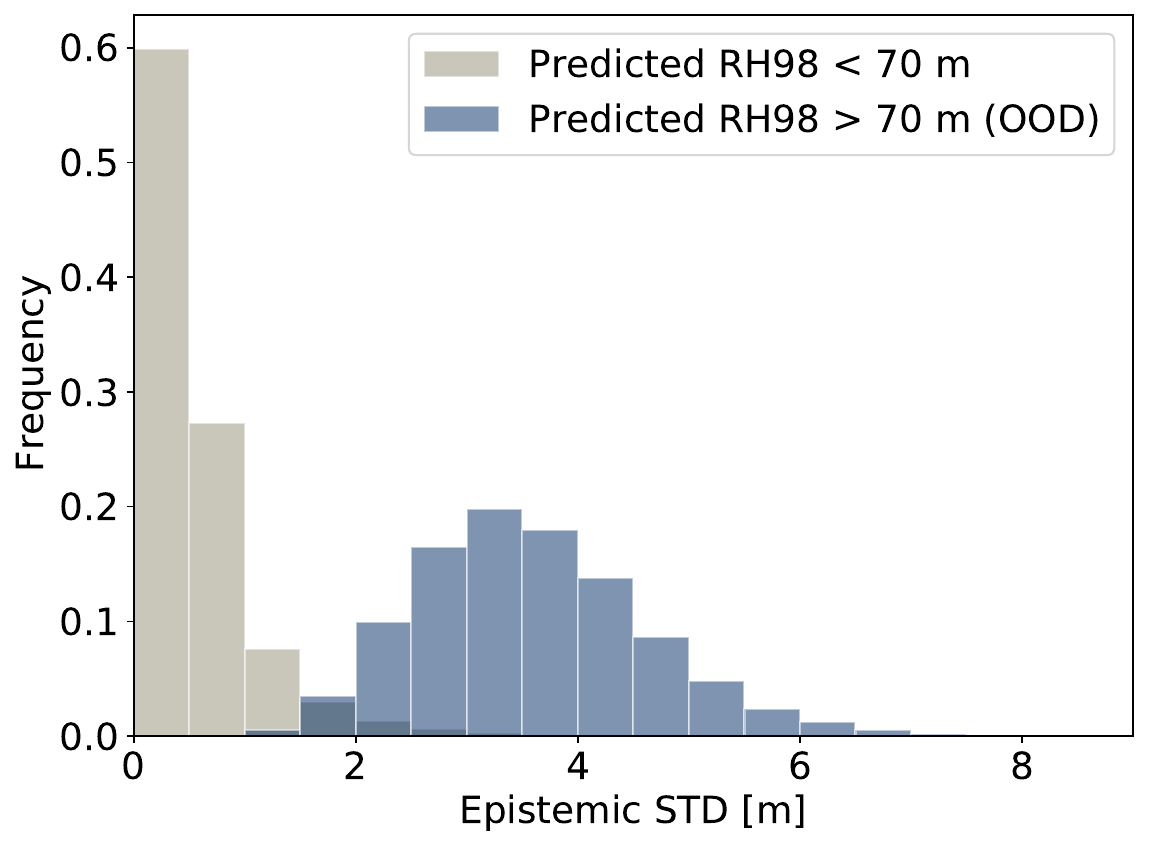} 
    \caption{Epistemic uncertainty for all predicted heights within the training data range (\textless70~m), and for all predictions exceeding the training data range (\textgreater70~m)\add{, i.e., out-of-distribution data (OOD)}.}%
    \label{fig:OOD_canopy_range}%
\end{figure*}

\newpage
\section{Estimating RH70 and ground elevation with the same model architecture}

\begin{figure*}[th]
    \centering
    \subfloat[]{{\includegraphics[width=0.4\textwidth]{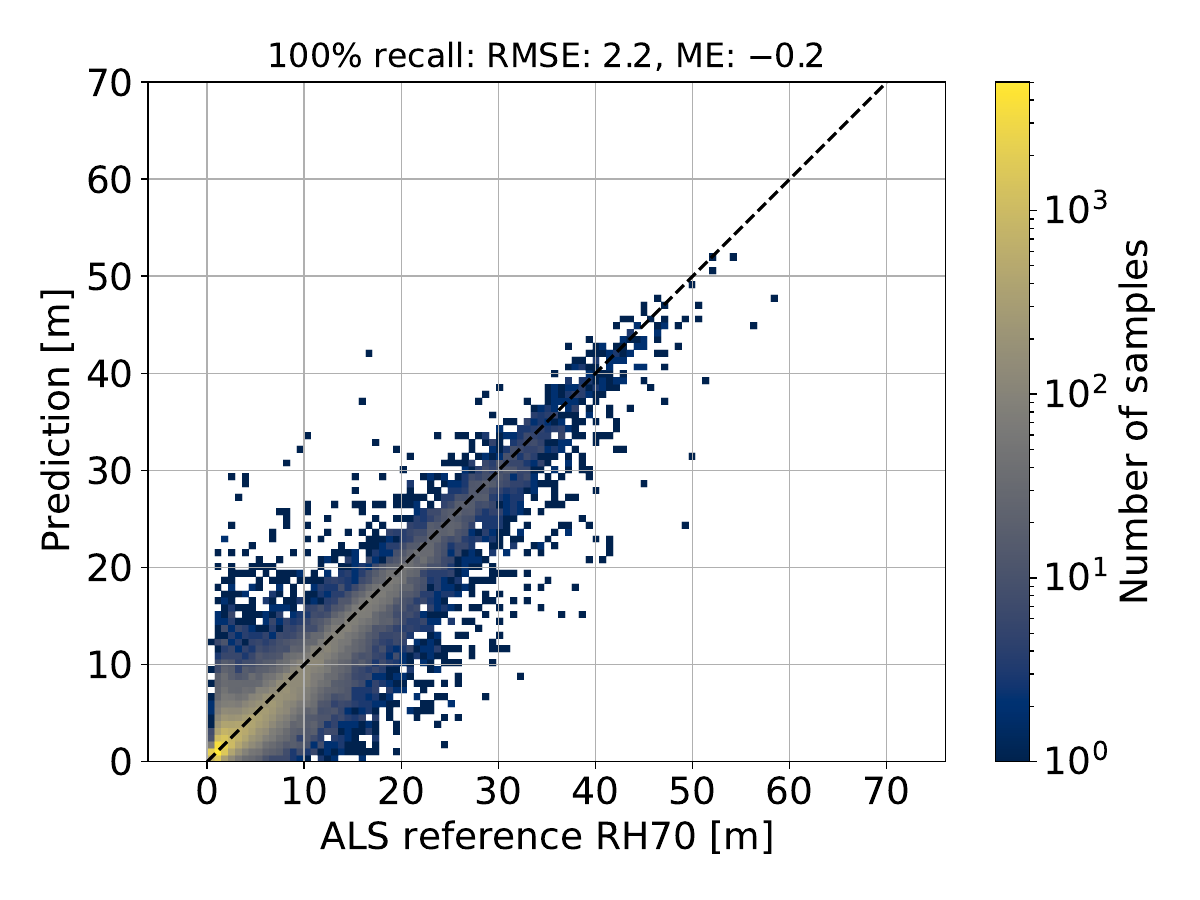} }}%
    \qquad
    \subfloat[]{{\includegraphics[width=0.4\textwidth]{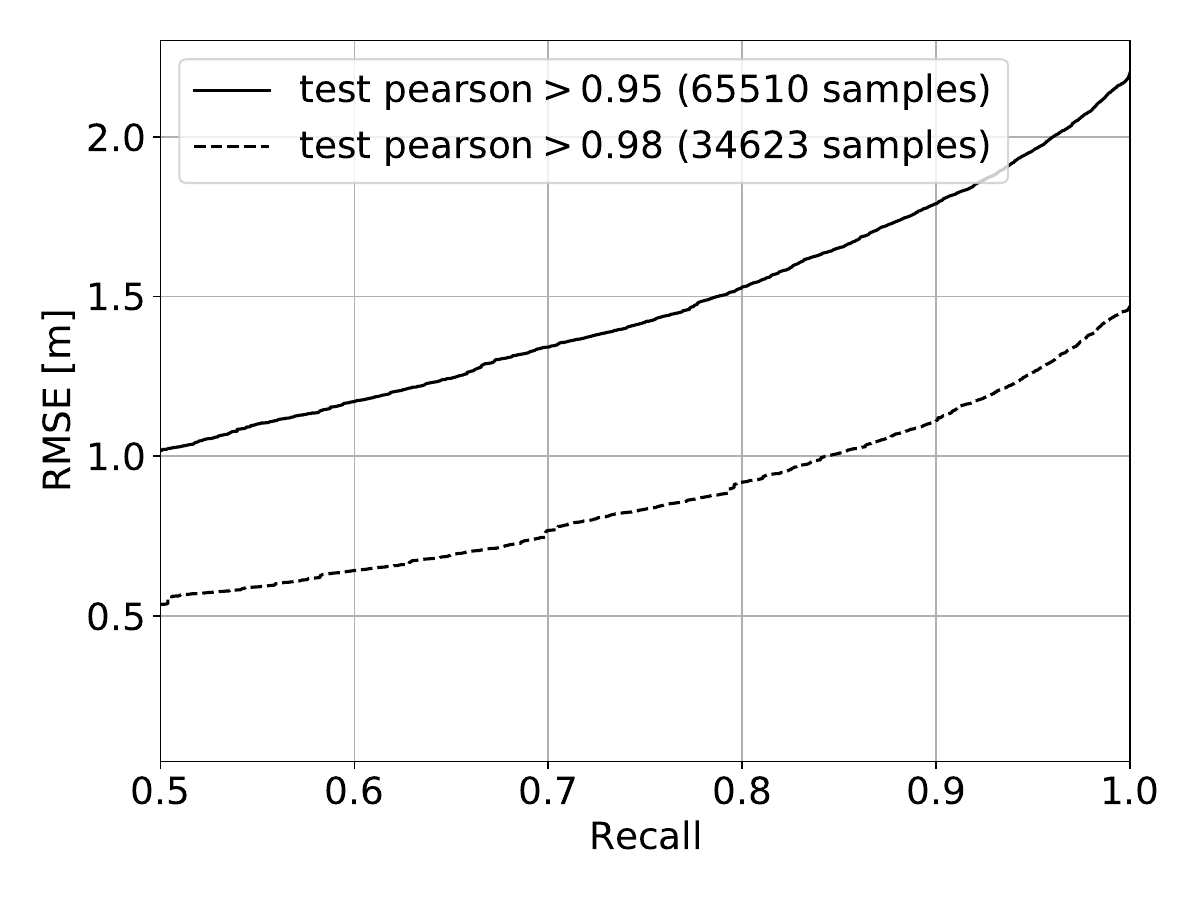} }}%
    \caption{Regression of RH70 from L1B GEDI full waveforms. (a) ALS reference data vs.\ CNN estimates, for all test samples. (b) Precision vs.\ recall when varying the threshold for the maximum predictive uncertainty. These are generated with the same adaptive thresholds as for RH98.}%
    \label{fig:regression_rh70}%
\end{figure*}

\begin{figure*}[th]
    \centering
    \subfloat[]{{\includegraphics[width=0.4\textwidth]{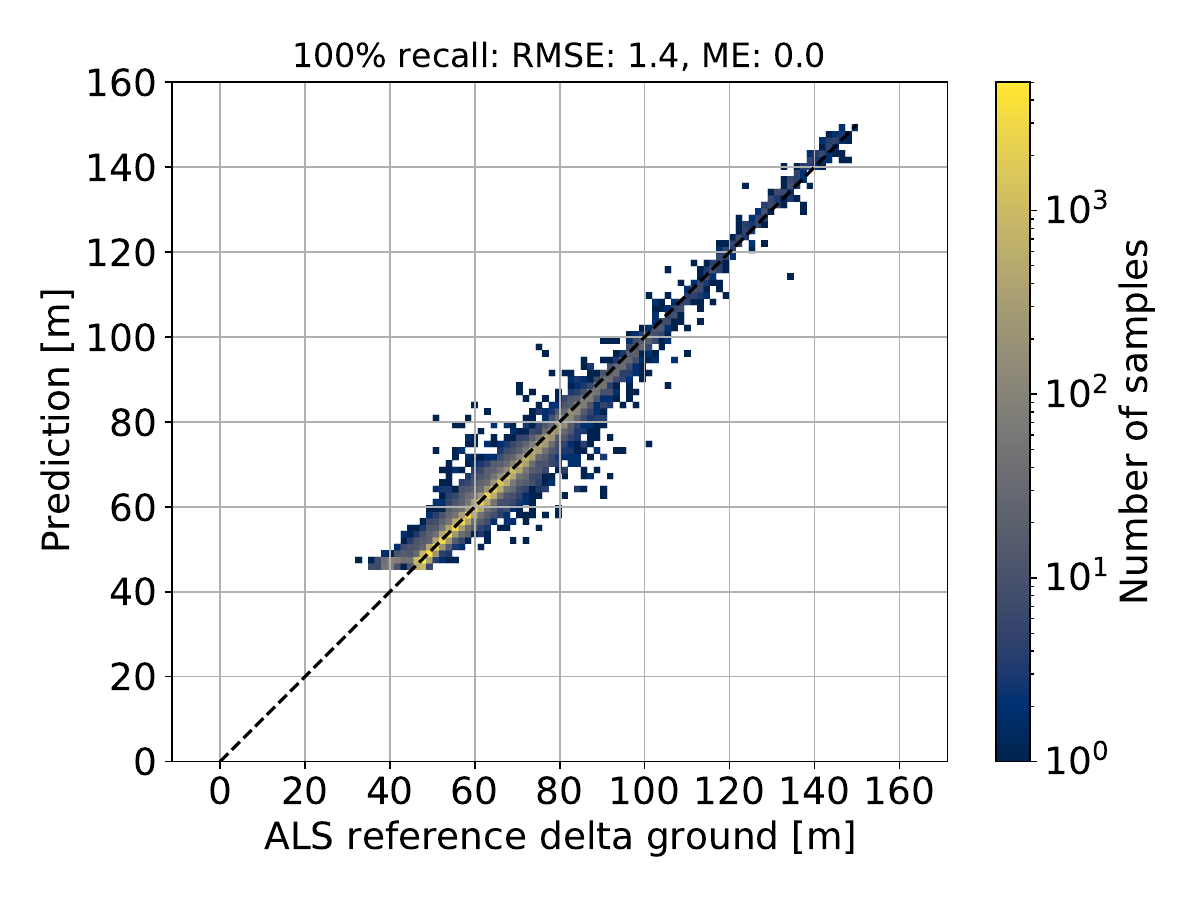} }}%
    \qquad
    \subfloat[]{{\includegraphics[width=0.4\textwidth]{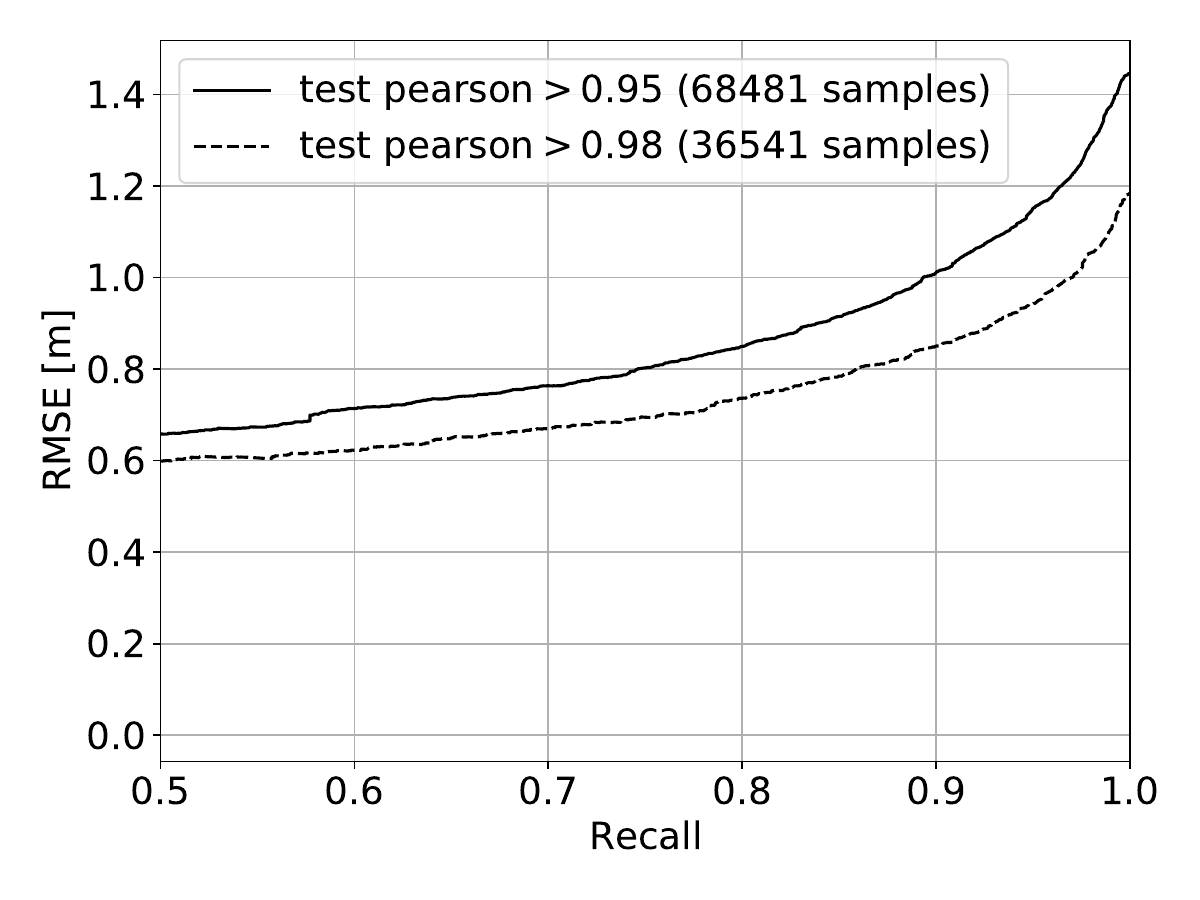} }}%
    \caption{Regression of ground elevation from L1B GEDI full waveforms. We train the model to predict the elevation difference between the first waveform return and the true ground elevation. I.e., the final ground elevation would be computed by adding the estimated value to the elevation that corresponds to the first return.
    (a) ALS reference data vs.\ the CNN estimates, for all test samples. (b) Precision vs.\ recall when varying the threshold for the maximum (absolute) predictive uncertainty.}%
    \label{fig:regression_ground}%
\end{figure*}



\end{document}